\documentclass[10pt,journal,compsoc]{IEEEtran}
\ifCLASSOPTIONcompsoc
  \usepackage[nocompress]{cite}
\else
  \usepackage{cite}
\fi
\ifCLASSINFOpdf
\else
\fi
\usepackage{amsmath}
\usepackage{amsfonts}
\usepackage{graphicx}
\usepackage{float}
\usepackage{booktabs}
\usepackage{multirow}
\usepackage[table,xcdraw]{xcolor}
\usepackage{overpic}
\usepackage{lineno}  %for line numbers
\usepackage{algorithm}
\usepackage{algorithmic}
\usepackage{multicol}

\begin{document}
\title{Joint Learning of Frequency and Spatial Domains for Dense Predictions}

\author{Shaocheng~JIA, \IEEEmembership{Student Member, IEEE}
        and Wei~YAO~\IEEEmembership{}% <-this % stops a space
\IEEEcompsocitemizethanks{\IEEEcompsocthanksitem This work has been submitted to the IEEE for possible publication. Copyright may be transferred without notice, after which this version may no longer be accessible. The work described in this paper was supported by the National Natural Science Foundation of China (Project No. 42171361). Corresponding author: W. YAO.
\IEEEcompsocthanksitem S. JIA is with the Department of Land Surveying and Geo-Informatics, The Hong Kong Polytechnic University, Hong Kong and the Department of Civil Engineering, The University of Hong Kong, Hong Kong;
W. YAO is with the Department of Land Surveying and Geo-Informatics, The Hong Kong Polytechnic University, Hong Kong and The Hong Kong Polytechnic University Shenzhen Research Institute, Shenzhen, China, E-mail: wei.hn.yao@polyu.edu.hk.
%E-mail: wei.hn.yao@polyu.edu.hk
}% <-this % stops an unwanted space
%\thanks{Manuscript received [month] [day], 2021; revised [month] [day], 2022.}
}

\IEEEtitleabstractindextext{%
\begin{abstract}
Current artificial neural networks mainly conduct the learning process in the spatial domain but neglect the frequency domain learning. However, the learning course performed in the frequency domain can be more efficient than that in the spatial domain. In this paper, we fully explore frequency domain learning and propose a joint learning paradigm of frequency and spatial domains. This paradigm can take full advantage of the preponderances of frequency learning and spatial learning; specifically, frequency and spatial domain learning can effectively capture global and local information, respectively. Exhaustive experiments on two dense prediction tasks, i.e., self-supervised depth estimation and semantic segmentation, demonstrate that the proposed joint learning paradigm can 1) achieve performance competitive to those of state-of-the-art methods in both depth estimation and semantic segmentation tasks, even without pretraining; and 2) significantly reduce the number of parameters compared to other state-of-the-art methods, which provides more chance to develop real-world applications. We hope that the proposed method can encourage more research in cross-domain learning.
\end{abstract}

}

\maketitle

\IEEEdisplaynontitleabstractindextext

% \fi
\IEEEpeerreviewmaketitle

\IEEEraisesectionheading{\section{Introduction}\label{sec:introduction}}

\IEEEPARstart{D}{eep} learning, as a heated research topic, has drawn much attention for several years, and it was also well studied and successfully applied to many fields, such as computer vision, speech recognition, and signal processing. More specifically, artificial neural networks (ANNs), especially a variety of convolutional neural networks (CNNs) \cite{goodfellow2016deep}, are dominated in achieving promising results on varying computer vision tasks.

CNNs, e.g., AlexNet \cite{krizhevsky2012imagenet} and ResNet \cite{he2016deep}, consist of stacked convolutional layers with different parameters, in which convolutional operation is the key to perform feature extraction and transformation. However, the convolutional operation has a limited receptive field by nature, which confines the convolution as a local operator in image processing. In reality, global information is on the critical path to many computer vision tasks, such as image classification, depth estimation, and semantic segmentation. To cope with the inherent shortcoming of the convolutional operator, a pyramid-like network architecture is commonly used in devising a convolutional network, gradually scaling down the feature map size for better extracting the global information. However, global information extraction is still challenging, despite the application of the special network architecture mentioned above.

Recently, self-attention-based Transformer \cite{Transformer} architecture provides a new perspective for feature extraction and transformation. Meanwhile, it is able to extract global information inherently. Nevertheless, Transformer \cite{Transformer} requires huge computational and storage resources, even though some improvements have been made (e.g., Linformer \cite{Linformer}).

Apart from the pyramid-like structure in CNNs and the self-attention Transformer model \cite{Transformer}, tremendous works have deployed conditional random fields (CRFs) and Markov random fields (MRFs) to capture the global information \cite{cao2017estimating}. These methods have difficulty in constructing an end-to-end network and being optimized. 

The aforementioned approaches all learn in the spatial domain, under which global information extraction depends on learning over all pixels in the image. Apparently, spatial domain learning is inefficient for global information extraction. By contrast, frequency-domain learning based on Fourier transform (FT) is more cost-effective for obtaining global information.

Fourier transform is widely used in signal processing. It transforms the temporal or spatial domain data to the frequency domain; then, the algorithms are applied to the frequency-domain data; finally, the processed frequency-domain data are converted back to the temporal or spatial data in light of the inverse Fourier transform (IFT). Notably, each frequency component obtained by FT is correlated with all temporal or spatial data in terms of the definition of FT. In image processing, each pixel obtained by discrete Fourier transform (DFT) is related to all pixels in spatial image. Therefore, all operations in frequency domain are globally spatial operations. Using operations in frequency domain we can easily obtain the global information of the input image. 

Several preliminary works in frequency domain learning have been conducted, especially for image classification \cite{frequencyclassification2017,frequencyclassification2020,frequencyclassification2021,fastFFTconv}. However, rare trials have been done in frequency domain learning for dense predictions, such as depth estimation and semantic segmentation; the dense prediction tasks, in general, simultaneously require global and local information, which is more challenging than the image classification task. Moreover, devising appropriate frequency learning operations and further identifying the suitable network architecture remain open. This study is inspired by these research gaps.

To sum up, the contributions of this paper are as follows:
\begin{itemize}
\item We propose an innovative linear frequency learning block (LFLBlock), which can effectively capture the global information in the frequency domain;

\item Utilizing the proposed LFLBlock and convolution operation we propose a joint frequency and spatial domains learning block, FSLBlock, to capture the global and local information simultaneously;

\item Leveraging the proposed FSLBlock we further propose a joint frequency and spatial domains learning network, FSLNet, to perform the dense predictions;

\item Applying the proposed network to different dense prediction tasks, i.e., self-supervised depth estimation and semantic segmentation, the competitive results and lightweight models demonstrate the superiority of the proposed method.  
\end{itemize}

The reminder of this paper is organized as follows. Section $2$ introduces related works. Section $3$ defines the problem. Section $4$ presents the proposed model. Section $5$ reports on detailed experiments. Section $6$ discusses the limitations of the proposed method. Section $7$ draws the conclusions.

\section{RELATED WORK}
In this section, literature regarding spatial domain learning and frequency domain learning are reviewed, respectively. 

\subsection{Spatial domain learning}
This paper focuses on dense prediction tasks in computer vision, e.g., self-supervised depth estimation and semantic segmentation, which differ from image classification, requiring both global and local feature information. However, pure CNNs are challenging to capture the global information of the input image, while naive Transformer \cite{Transformer} has difficulty in capturing the local information. Therefore, numerous works have been done in improving feature extraction and transformation.

In CNNs, there are three types of strategies to enhance the global information: 1) cross-scale fusion, which integrates multi-scale features to gather various levels of information, such as FCN \cite{long2015fully}, FPN \cite{lin2017feature}, and HRNet \cite{wang2020deep}; 2) non-local neural networks, which aims at devising the non-local operations to capture the global features \cite{wang2018non, zhu2019asymmetric, huang2019ccnet, chi2020non, luo2016understanding}, e.g., linking the arbitrary distant neurons; and 3) external auxiliary, which mainly uses additional structures and tools to help extract the global information, such as CRFs and MRFs \cite{eigen2015predicting,li2015depth,liu2015deep,mousavian2016joint,xu2017multi,xu2018structured,karschdepth,saxena20083,shen2017semantic,liu2015crf}.

In Transformer-based models, a large amount of works use CNNs as either the encoder \cite{T4C1, T4C2, T4C3, T4C4,yang2021transformers} or decoder \cite{C4T1, C4T2, C4T3, C4T4,ranftl2021vision} to capture the fine details. Moreover, the soft split strategy \cite{C4T4,mypaper2} and pyramid-like structure \cite{wang2021pyramid, graham2021levit} are also proposed to efficiently capture the local features.

Summarily, spatial domain learning, especially CNNs, has advantages in extracting the fine features. Although Transformer-based models can capture global information, they require huge computational and storage resources. The proposed framework takes both efficiency and effectiveness into account.

\subsection{Frequency domain learning}
Most deep learning methods extract features and perform feature transformations solely in the spatial domain. Spatial domain learning is intuitively understandable because it is consistent with the human's observations. However, current frameworks proposed for spatial domain learning have limitations in either global or local information extraction, e.g., CNNs and Transformer-based models \cite{Transformer}. In another perspective, frequency-domain learning which transforms the input data to the frequency domain, by nature, can capture global contextual information.

Using DFT, a few works attempted to design frequency learning frameworks or operations for conducting image classification. Specifically, Stuchi et al.  \cite{frequencyclassification2017} first utilized Fourier analysis to extract the discriminative feature for improving face liveness detection; Further, Stuchi et al. \cite{frequencyclassification2020} proposed a frequency learning framework, which performs the learning process over different sizes of patches for capturing the multi-level features; Chi et al. \cite{fastFFTconv} proposed Fourier unit and local Fourier unit to extract global and local features; and Watanabe et al. \cite{frequencyclassification2021} proposed 2SReLU layer, which can preserve the high-frequency components in deep networks.

The literature presented above has demonstrated the effectiveness of frequency-domain learning in image classification. However, to the best of our knowledge, no trials apply frequency-domain learning to dense prediction tasks. Differing from image classification, dense prediction tasks require capturing more fine details except for the global contextual information, which is more complicated.

Therefore, we propose joint frequency and spatial domains learning, taking full advantage of their superiority in extracting global and local features, respectively.

\section{PROBLEM SETUP}
It is critical but challenging to preserve global and local information during the learning process, especially for high-dimensional data like images. Let $F_{in} \in \mathbb{R}^{b \times c \times h \times w}$ represent the input images or features of a learning block, wherein $b$, $c$, $h$, and $w$ represent the batch size, the number of channels, the height of the input features, and the width of the input features, respectively. Similarly, let $F_{out} \in \mathbb{R}^{b \times c^{'} \times h^{'} \times w^{'}}$ be the output features of a learning block, wherein $c^{'}$, $h^{'}$, and $w^{'}$ represent the target number of channel, the target height of the output features, and the target width of the output features, respectively.

Given $F_{in}$ and $F_{out}$, the primary task is to devise a learning block that can extract both global and local information from the input; we can mathematically define it as: $\Psi: F_{in} \to F_{out}$, where $\Psi$ represents the unknown leaning block. Moreover, the subsequent task is to design a learning network that can perform the dense prediction tasks in an end-to-end manner; which can be mathematically defined as: $G=f(\Psi_1, \Psi_2, ..., \Psi_n)$, where $G$ and $n$ represent the unknown learning network and the number of learning blocks. This paper is to propose innovative $\Psi$ and $G$. Then, we apply the proposed network to dense prediction tasks to demonstrate its efficiency and effectiveness.

\section{METHOD}
In this section, the linear frequency learning block (LFLBlock), the joint frequency and spatial learning block (FSLBlock), and the joint frequency and spatial domains learning network (FSLNet) are introduced, respectively.

\subsection{Linear frequency learning block}
Extracting the global information from the input image is extremely challenging for CNNs, due to the limited receptive field. Obviously, global information is correlated with all pixels in the image, which indicates that all pixels should be taken into account to capture the global information. In which case, a spatial operation kernel having the size identical to that of the input image is required, however, which is not achievable in reality in the interest of limited computational and storage resources.

Let us consider the problem mentioned above in the frequency domain instead of the spatial domain. Using DFT, the spatial image can be readily transformed to the frequency domain, as shown in Eq. \ref{dft}.
\begin{equation}
\label{dft}
\begin{aligned}
    F(u,v) &= \sum_{x=0}^{H-1}\sum_{y=0}^{W-1}I(x,y)e^{-j2\pi(\frac{ux}{H}+\frac{vy}{W})} 
    \\u &=0,1,2,...,H-1
    \\v &=0,1,2,...,W-1
\end{aligned},
\end{equation}
wherein $F \in \mathbb{R}^{H \times W \times c_{in}}$, $I \in \mathbb{R}^{H \times W \times c_{in}}$, $j$, $H$, and $W$ represent the transformed frequency data, the spatial image, the imaginary unit, the height of the input image, and the width of the input image, respectively. Note that $F$ is a complex matrix. To conveniently perform the learning process, we rewrite the frequency-domain data $F$ as follows (Eq. \ref{concat}):
\begin{equation}
\label{concat}
    F_{1} = Concat(Re(F), Im(F)),
\end{equation}
where $F_{1} \in \mathbb{R}^{H \times W \times 2c_{in}}$, $Concat(\cdot)$, $Re(\cdot)$, and $Im(\cdot)$ represent the rewritten frequency data, the concat operation by channel, the function of obtaining the real part, and the function of obtaining the imaginary part, respectively.

Since each frequency pixel of $F_{1}$ represents different frequency components and each frequency component is deduced by all spatial pixels, all operations in the frequency domain are global operations for the spatial domain. More specifically, the global information extraction in the spatial domain can be substituted with the local operation in the frequency domain alternatively.

Therefore, we propose linear frequency learning block (LFLBlock), applying it to all frequency components to perform the frequency domain learning, i.e. (Eq. \ref{slo}),
\begin{equation}
\label{slo}
    F_{2} =  F_{1}w,
\end{equation}
wherein $F_{2} \in \mathbb{R}^{H \times W \times 2c_{out}}$ and $w \in \mathbb{R}^{2c_{in} \times 2c_{out}}$ represent the learned feature in frequency domain and the learnable parameters.

After linear frequency learning, we apply activation function and normalization, i.e. (Eq. \ref{actnorm}),
\begin{equation}
\label{actnorm}
    F_{3} =  BN(SiLU(F_{2})),
\end{equation}
wherein $SiLU(\cdot)$ and $BN(\cdot)$ represent the SiLU activation function (i.e., swish function) and the batch normalization. Note that the number of linear frequency learning layers can be specified according to different tasks in practice.

For transforming the learned frequency-domain feature to spatial domain feature, we first convert the learned real matrix to the complex matrix in terms of Eq. \ref{putback},
\begin{equation}
\label{putback}
    F_{4} =  F_{3}^{1} + jF_{3}^{2},
\end{equation}
wherein $F_{3}^{1}$ and $F_{3}^{2}$ represent the first and second $c_{out}-$channel data of $F_{3}$, respectively; $F_{4} \in \mathbb{R}^{H \times W \times c_{out}}$ is the learned complex matrix.

Finally, inverse discrete Fourier transform (IDFT) is used to convert the learned feature in the frequency domain to spatial feature, i.e. (Eq. \ref{idft}),
\begin{equation}
\label{idft}
\begin{aligned}
    I_{1}(x,y) &= \sum_{u=0}^{H-1}\sum_{v=0}^{W-1}F_{4}(u,v)e^{j2\pi(\frac{ux}{H}+\frac{vy}{W})} 
    \\x &=0,1,2,...,H-1
    \\y &=0,1,2,...,W-1
\end{aligned},
\end{equation}
where $I_{1} \in \mathbb{R}^{H \times W \times c_{out}}$ is the learned spatial feature from the linear frequency learning block.

\begin{algorithm} 
	\caption{PyTorch-like implementation of Linear frequency learning block.} 
	\label{lflblock} 
	\begin{algorithmic}[1]
	    \STATE \textbf{Input}: $x \in \mathbb{R}^{b \times in\_c \times h \times w}$, $in\_c \in \mathbb{R}^{1}$, $out\_c \in \mathbb{R}^{1}$, $layer\_num \in \mathbb{R}^{1}$ 
		\STATE Initialization: L $\gets$ []
		\FOR{$i$ in range($layer\_num$)}
		\IF{$i = 0$}
		\STATE L.append(Conv2d($2 \times in\_c$, $2 \times out\_channel $, $1$, bias=False))
		\STATE L.append(SiLU(True))	
		\STATE L.append(BatchNorm2d($2 \times out\_c$))	\ELSE
		\STATE L.append(Conv2d($2 \times out\_c$, $out\_c$, $1$, bias=False))
		\STATE L.append(SiLU(True))	
		\STATE L.append(BatchNorm2d($2 \times out\_c$))	\ENDIF
		
		\ENDFOR
		
		\STATE $x$ $\gets$ fft.rfft($x$, dim=($2$, $3$, norm='ortho'))
		\STATE $R$ $\gets$ real($x$)
		\STATE $I$ $\gets$ imag($x$)
		\STATE $x$ $\gets$ cat($R$, $I$, dim=$1$)
		
		\FOR{layer in L}
		\STATE $x$ $\gets$ layer($x$)
		\ENDFOR
		
		\STATE $R$ $\gets$ $x$[:,:$out\_c$,:,:]
		\STATE $I$ $\gets$ $x$[:,$out\_c$:,:,:]
		\STATE $x$ $\gets$ $R + (j \times I)$
		\STATE $x$ $\gets$ irfftn($x$, dim=($2$, $3$), norm='ortho')
		
		\STATE \textbf{Output}: $x \in \mathbb{R}^{b \times out\_{c} \times h \times w}$
	\end{algorithmic} 
\end{algorithm}

\begin{algorithm*} 
	\caption{PyTorch-like implementation of convolutional neural network block (CNNBlock).} 
	
	\label{cnn} 
	\begin{multicols}{2}
	
	\begin{algorithmic}[1]
	    \STATE \textbf{Input}: $x \in \mathbb{R}^{b \times in\_c \times h \times w}$, $in\_c \in \mathbb{R}^{1}$, $out\_c \in \mathbb{R}^{1}$, $layer\_num \in \mathbb{R}^{1}$, $dim \in \mathbb{R}^{1}$, $bottleneck \in {0, 1}$

		\STATE Initialization: L $\gets$ []
		
	    \STATE \textbf{Function} CNN($in\_c, out\_c, k$)
	    \STATE  \quad layers $\gets$ []
	    \STATE  \quad layers.append(Conv2d($in\_c, out\_c, k, padding = k//2, padding\_mode = 'reflect', bias = False$))
	    \STATE  \quad layers.append(SiLU(True))
	    \STATE  \quad layers.append(BatchNorm2d($out\_c$))
	    \STATE  \quad \textbf{Return:} nn.Sequential(layers)

		\IF{$bottleneck=1$}
		\IF{$in\_c \leq dim$ and $out\_c \leq dim$}
		\FOR{$i$ in range($layer\_num$)}
		\IF{$i = 0$}
        \STATE L.append(CNN($in\_c$, $out\_c$))
        \ELSE
        \STATE L.append(CNN($out\_c$, $out\_c$))        
        \ENDIF
		\ENDFOR
		
		\ELSIF{$in\_c > dim$ and $out\_c \leq dim$}
		\FOR{$i$ in range($layer\_num$)}
		\IF{$i = 0$}
        \STATE L.append(CNN($in\_c$, $dim$, $k=1$)) 
        \STATE L.append(CNN($dim$, $out\_c$))   
        \ELSE
        \STATE L.append(CNN($out\_c$, $out\_c$)) 
        \ENDIF
		\ENDFOR		

		\ELSIF{$in\_c > dim$ and $out\_c > dim$}
		\FOR{$i$ in range($layer\_num$}
		\IF{$i = 0$}
        \STATE L.append(CNN($in\_c$, $dim$, $k=1$)) 
        \STATE L.append(CNN($dim$, $dim$)) 
        \IF{$i=layer\_num-1$}
        \STATE L.append(CNN($dim$, $out\_c$, $k=1$)) 
        \ENDIF
        \ELSE
        \IF{$i \ne layer\_dim-1$}
        \STATE L.append(CNN($dim$, $dim$)) 
        \ELSE
        \STATE L.append(CNN($dim$, $dim$)) 
        \STATE L.append(CNN($dim$, $out\_c$, $k=1$)) 
        \ENDIF
        \ENDIF
		\ENDFOR
		
		\ELSIF{$in\_c \leq dim$ and $out\_c > dim$}
		\FOR{$i$ in range($layer\_num$}
		\IF{$i = 0$}
        \STATE L.append(CNN($in\_c$, $dim$)) 
        \IF{$i=layer\_num-1$}
        \STATE L.append(CNN($dim$, $out\_c$, $k=1$)) 
        \ENDIF
        \ELSE
        \IF{$i \ne layer\_dim-1$}
        \STATE L.append(CNN($dim$, $dim$)) 
        \ELSE
        \STATE L.append(CNN($dim$, $dim$)) 
        \STATE L.append(CNN($dim$, $out\_c$, $k=1$)) 
        \ENDIF
        \ENDIF
		\ENDFOR	
		
		\ENDIF

		\ELSE
		\FOR{$i$ in range($layer\_num$}
        \IF{$i = 0$}
        \STATE L.append(CNN($in\_c$, $out\_c$)) 
        \ELSE
        \STATE L.append(CNN($out\_c$, $out\_c$)) 
        \ENDIF
		\ENDFOR			
		\ENDIF
		
		\FOR{layer in L}
		\STATE $x \gets$ layer($x$)
		\ENDFOR

		\STATE \textbf{Output}: $x \in \mathbb{R}^{b \times out\_{c} \times h \times w}$
	\end{algorithmic} 
	\end{multicols}
\end{algorithm*}

During implementation, the DFT of a real-valued two-dimensional (2D) matrix is conjugate symmetric, which can help reduce the computational cost. Moreover, the proposed LFLBlock is readily implemented using a convolutional operation with the kernel size of $1\times1$. PyTorch-like implementation of the proposed LFLBlock is presented in Algorithm \ref{lflblock}.

The proposed LFLBlock performs the learning process in the frequency domain instead of the spatial domain. Thus, it is naturally able to capture the global information of the input image. However, we also can find that the presented learning paradigm lacks anchor points since each frequency component is responsible for all spatial pixels. Thus, pure LFLBlock is challenging to capture the fine details of the input image. We will cope with this issue in the next subsection.

\subsection{Joint frequency and spatial learning block}
The proposed linear frequency learning block (LFLBlock) can effectively capture the global information of the input image but has limitations in extracting the fine details of the input image. In this subsection, the proposed LFLBlock and convolutional block are integrated to extract the comprehensive features, i.e., including global and local information simultaneously.

\begin{figure}[htbp]
    \centering
    \includegraphics[scale=0.1]{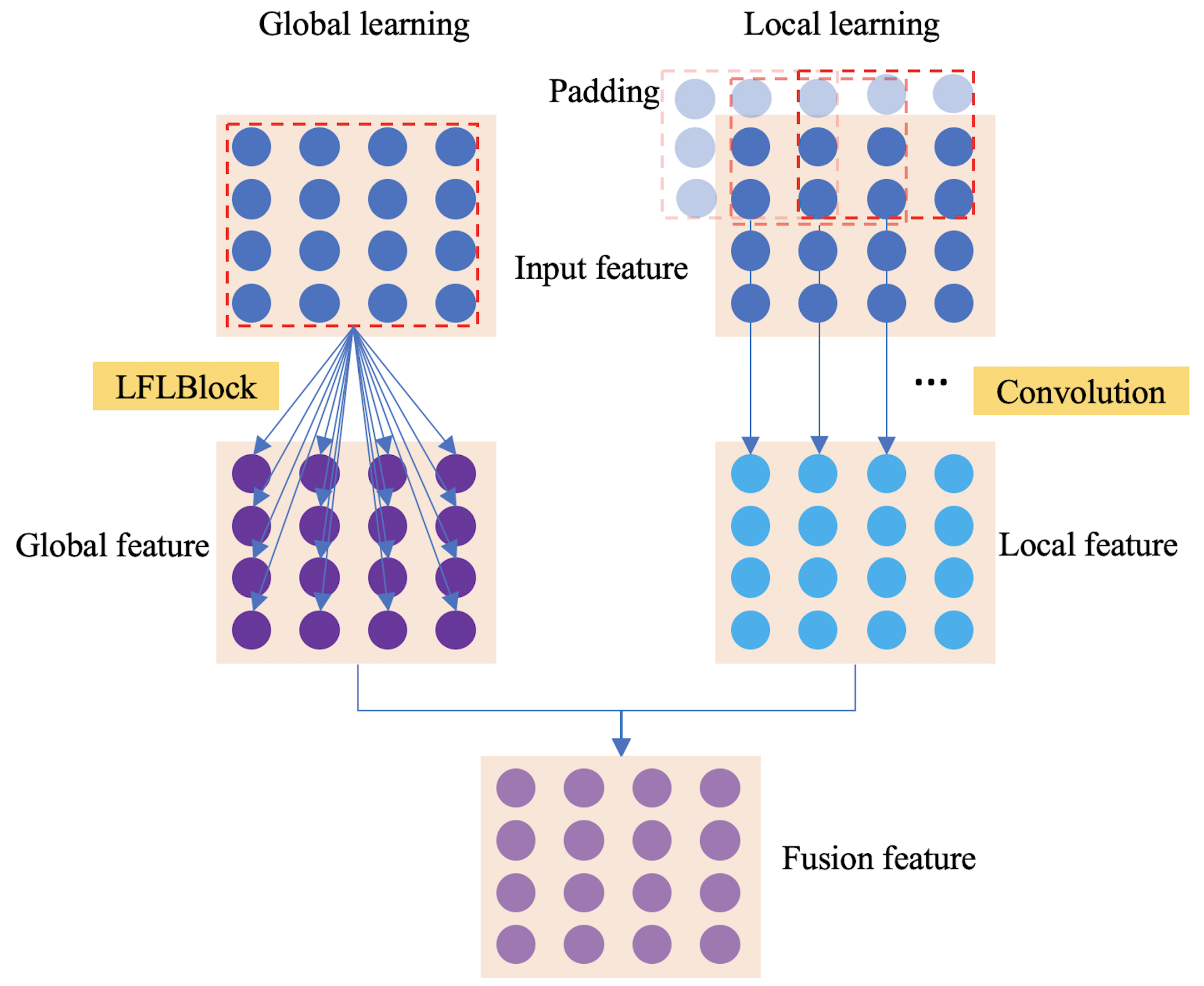}
    \caption{Illustration of joint frequency and spatial domains learning block (FSLBlock).}
    \label{FSLBlock}
\end{figure}

Figure \ref{FSLBlock} illustrates the proposed joint frequency and spatial domains learning block (FSLBlock), which is straightforward but effective. Specifically, there are two computational streams: global and local feature extraction. Following the notations stated in Section 4.1, let the input feature/image and the learned global feature be $I \in \mathbb{R}^{H \times W \times c_{in}}$ and $I_{1} \in \mathbb{R}^{H \times W \times c_{out}}$, respectively, thus, the global feature extraction can be formulated as Eq. \ref{global}:
\begin{equation}
\label{global}
    I_{1} = LFLBlock(I),
\end{equation}
where $LFLBlock(\cdot)$ represents the proposed LFLBlock in Section 4.1.

In parallel, CNNs are used to extract the local features from the input image, which can be described as Eq. \ref{local}:
\begin{equation}
\label{local}
    I_{2} = CNNs(I),
\end{equation}
where $I_{2} \in \mathbb{R}^{H \times W \times c_{out}}$ is the learned local features; $CNNs$ represents the convolutional blocks, each of which consists of a convolution layer with a $3 \times 3$ kernel (without bias), a SiLU activation layer, and a batch normalization layer. Being consistent with the linear frequency learning block, the number of convolution layers can be arbitrarily specified. Moreover, the bottleneck-like structure is adopted for CNNs to reduce the computational and storage costs when inputting 
or outputting high-dimensional data. PyTorch-like implementation of the used CNN block is shown in Algorithm \ref{cnn}.
%\begin{figure}[htbp]
%    \centering
%    \includegraphics[scale=0.19]{3.jpg}
%    \caption{PyTorch-like implementation of the used CNN block.}
%    \label{cnn}
%\end{figure}

Then, the global feature from LFLBlock and the local feature from the CNN blocks are integrated by a single CNN layer with the kernel size of $3 \times 3$ (without bias), the input channels $2c_{out}$, and the output channels $c_{out}$, which can be described as Eq. \ref{compression}:
\begin{equation}
\label{compression}
    I_{3} = BN(SiLU(Conv(Concat(I_{1}, I_{2})))),
\end{equation}
wherein $I_{3} \in \mathbb{R}^{H \times W \times c_{out}}$ is the learned comprehensive features.

The proposed joint frequency and spatial learning block (FSLBlock) is quite straightforward and easily implemented with the aid of the current convolutional operation. It also meets the intuition that integrating global and local information can obtain comprehensive information. From the perspective of information theory, the proposed FSLBlock can preserve the global information, thereby reducing the information loss along the course of feature extraction and transformation.

\subsection{Joint frequency and spatial learning network}
Using the proposed FSLBlock, a joint frequency and spatial learning network (FSLNet) is presented in this subsection.
\begin{figure*}[htbp]
    \centering
    \includegraphics[width=\textwidth]{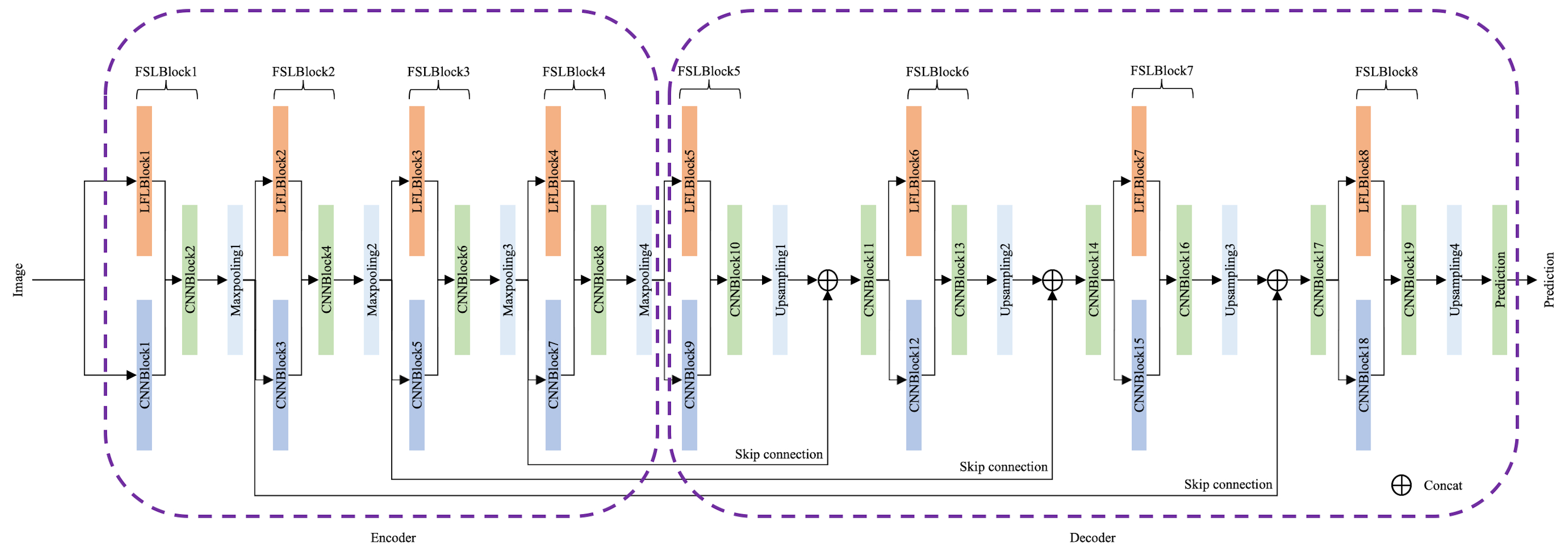}
    \caption{Architecture of the proposed joint frequency and spatial learning network (FSLNet).}
    \label{fsl}
\end{figure*}

Figure \ref{fsl} shows the architecture of the proposed FSLNet. Overall, the proposed FSLNet is designed in an encoder-decoder manner. Both encoder and decoder parts contain four stages. Specifically, each encoding stage is an FSLBlock, and a $MaxPool$ layer with the kernel size of $3 \times 3$, the stride of $2$, and the padding of $1$ is used between two stages; in decoding stages, the $MaxPool$ layer is substituted with an $Upsampling$ layer with scaling factor $2$ to gradually recover the original resolution. 

\begin{table}[htp]
\centering
\caption{Configurations of the proposed FSLNet. In\_ch, out\_ch, and BotN represent input channels, output channels, and bottleneck. c is the basic number of channels. Y and N represent using and ignoring bottleneck, respectively. Type column indicates the property of the corresponding computational block; specifically, Global, Local, and Fusion represent global information extraction, local information extraction, and fusing global and local features, respectively.}
\begin{tabular}{lccccc}
\toprule
Block & \# of layers & Type & In\_ch & Out\_ch & BotN \\ \midrule
LFLBlock1 & 2 & Global & 3 & c & -  \\
CNNBlock1 & 4 & Local & 3 & c & Y \\
CNNBlock2 & 1 & Fusion & 2c & c & N \\
LFLBlock2 & 2 & Global & c & 2c & -  \\
CNNBlock3 & 4 & Local & c & 2c & Y \\
CNNBlock4 & 1 & Fusion & 4c & 2c & N \\
LFLBlock3 & 2 & Global & 2c & 4c & -  \\
CNNBlock5 & 4 & Local & 2c & 4c & Y \\
CNNBlock6 & 1 & Fusion & 8c & 4c & N \\
LFLBlock4 & 2 & Global & 4c & 8c & -  \\
CNNBlock7 & 4 & Local & 4c & 8c & Y \\
CNNBlock8 & 1 & Fusion & 16c & 8c & N \\
LFLBlock5 & 2 & Global & 8c & 4c & -  \\
CNNBlock9 & 4 & Local & 8c & 4c & Y \\
CNNBlock10 & 1 & Fusion & 8c & 4c & N \\
CNNBlock11 & 1 & Fusion & 8c & 4c & N \\
LFLBlock6 & 2 & Global & 4c & 2c & -  \\
CNNBlock12 & 4 & Local & 4c & 2c & Y \\
CNNBlock13 & 1 & Fusion & 4c & 2c & N \\
CNNBlock14 & 1 & Fusion & 4c & 2c & N \\
LFLBlock7 & 2 & Global & 2c & c & -  \\
CNNBlock15 & 4 & Local & 2c & c & Y \\
CNNBlock16 & 1 & Fusion & 2c & c & N \\
CNNBlock17 & 1 & Fusion & 2c & c & N \\
LFLBlock8 & 2 & Global & c & c & -  \\
CNNBlock18 & 4 & Local & c & c & Y \\
CNNBlock19 & 1 & Fusion & 2c & c & N \\

\bottomrule
\end{tabular}
\label{parameters}
\end{table}

Moreover, three skip connections are used to fuse the multi-level features. Finally, a $Prediction$ block is used to predict the outputs. In this paper, we use three convolution layers as the prediction block to refine the results; detailed configurations regarding the Prediction block are introduced in Section 5. Detailed network configurations are presented in Table \ref{parameters}. Adjusting $c$ we can obtain models of having different representation capabilities. Specifically, we present two models, FSLNet-S and FSLNet-L, in which $c$ is equal to $16$ and $32$, respectively.  

Although we focus on dense prediction tasks in this paper, the proposed FSLNet also can be used to conduct other sparse prediction tasks, via dropping or modifying the decoder, e.g., the ego-motion estimation in self-supervised depth estimation. In the next section, the proposed FSLNet is applied to two dense prediction tasks, i.e., self-supervised depth estimation and semantic segmentation, to demonstrate its effectiveness and superiority.

\section{EXPERIMENT}

In this section, self-supervised depth estimation, semantic segmentation, experimental results, and ablation studies are introduced, respectively. All experiments are implemented with PyTorch $1.9.1$ on a single TITAN RTX card.

\subsection{Self-supervised depth estimation}

\subsubsection{Task description}
Monocular depth estimation is a low-level computer vision task and plays an important role in many fields, such as autonomous driving, three-dimensional (3D) reconstruction, and robotics. Roughly speaking, there are two types of paradigms in monocular depth estimation: supervised depth estimation \cite{saxena20083, eigen2015predicting, mypaper2020} and self-supervised depth estimation \cite{zhou2017unsupervised, godard2019digging, mypaper2021, mypaper2}. In 
these two approaches, self-supervised depth estimation has drawn much attention in recent years, because it requires nothing but image sequence during training, thereby being able to conveniently apply it to different scenarios.

Self-supervised depth estimation system consists of two networks, i.e., depth estimation network, $G_{D}: I_{2} \in \mathbb{R}^{H \times W \times C} \to D \in \mathbb{R}^{H \times W}$, and pose estimation network, $G_{P}: I_{i} \in \mathbb{R}^{H \times W \times C},i=1,2,3 \to P \in \mathbb{R}^{2 \times 6}$, which are jointly trained during the training phase but can separately work at inference, wherein $I_{i}$, $D$, and $P$ represent the source images, the estimated depth map, and the estimated pose vectors. Note that $I_{i}, i=1,2,3$ are consecutive frames and $I_{2}$ is the middle frame. Given the depth map of the target image $I_{2}$ and the relative movements between the target image and the reference images (i.e., $I_{1}$ and $I_{3}$), we can use the reconstruction loss of the view synthesis to train the system.

\begin{table*}[h]
\caption{Depth estimation results on the KITTI dataset \cite{geiger2013vision}. GT, PT, SC, TC, M, and G represent ground truth, pretraining, space complexity, time complexity, megabyte, and GFLOPs respectively. - represents that the situation is unclear. Res. represents resolution. The best performances and our models are marked \textbf{bold}. The second-best performances in the second and the third cells are \underline{underlined}. * we ran Monodepth2 codes in our platform to generate the results without pretraining at a resolution of $416\times128$; please note that we did not change any parameters except for pretraining.}
\small
\centering
\begin{tabular}{@{}lccccccccccc@{}}
%\toprule
\toprule
{\color[HTML]{000000} } & {\color[HTML]{000000} } & {\color[HTML]{000000} } & {\color[HTML]{000000} } & {\color[HTML]{000000} } & \multicolumn{4}{c}{{\color[HTML]{000000} Errors $\downarrow$}} & \multicolumn{3}{c}{{\color[HTML]{000000} Errors $\uparrow$}} \\ \cline{6-12} %\cmidrule(l){6-12} 
\multirow{-2}{*}{{\color[HTML]{000000} Methods}} & \multirow{-2}{*}{{\color[HTML]{000000} GT?}} & \multirow{-2}{*}{{\color[HTML]{000000} PT?}} & \multirow{-2}{*}{{\color[HTML]{000000} SC?}} & \multirow{-2}{*}{{\color[HTML]{000000} TC?}} & {\color[HTML]{000000} AbsRel} & {\color[HTML]{000000} SqRel} & {\color[HTML]{000000} RMS} & {\color[HTML]{000000} RMSlog} & {\color[HTML]{000000} $<1.25$} & {\color[HTML]{000000} $<1.25^{2}$} & {\color[HTML]{000000} $<1.25^{3}$} \\ \hline %\midrule
{\color[HTML]{000000} Eigen et al., coarse   \cite{eigen2014depth}} & \multicolumn{1}{c}{{\color[HTML]{000000} $\surd$}} & {\color[HTML]{000000} -} & {\color[HTML]{000000} -} & {\color[HTML]{000000} -} & {\color[HTML]{000000} 0.214} & {\color[HTML]{000000} 1.605} & {\color[HTML]{000000} 6.563} & {\color[HTML]{000000} 0.292} & {\color[HTML]{000000} 0.673} & {\color[HTML]{000000} 0.884} & {\color[HTML]{000000} 0.957} \\
{\color[HTML]{000000} Eigen et al., fine \cite{eigen2014depth}} & \multicolumn{1}{c}{{\color[HTML]{000000} $\surd$}} & {\color[HTML]{000000} -} & {\color[HTML]{000000} -} & {\color[HTML]{000000} -} & {\color[HTML]{000000} 0.203} & {\color[HTML]{000000} 1.548} & {\color[HTML]{000000} 6.307} & {\color[HTML]{000000} 0.282} & {\color[HTML]{000000} 0.702} & {\color[HTML]{000000} 0.890} & {\color[HTML]{000000} 0.958} \\
{\color[HTML]{000000} Liu et al. \cite{liu2015learning}} & \multicolumn{1}{c}{{\color[HTML]{000000} $\surd$}} & {\color[HTML]{000000} -} & {\color[HTML]{000000} -} & {\color[HTML]{000000} -} & {\color[HTML]{000000} 0.202} & {\color[HTML]{000000} 1.614} & {\color[HTML]{000000} 6.523} & {\color[HTML]{000000} 0.275} & {\color[HTML]{000000} 0.678} & {\color[HTML]{000000} 0.895} & {\color[HTML]{000000} 0.965} \\
{\color[HTML]{000000} Kuznietsov et al. \cite{kuznietsov2017semi}} & \multicolumn{1}{c}{{\color[HTML]{000000} $\surd$}} & {\color[HTML]{000000} $\surd$} & {\color[HTML]{000000} -} & {\color[HTML]{000000} -} & {\color[HTML]{000000} 0.113} & {\color[HTML]{000000} 0.741} & {\color[HTML]{000000} 4.621} & {\color[HTML]{000000} 0.189} & {\color[HTML]{000000} 0.862} & {\color[HTML]{000000} 0.960} & {\color[HTML]{000000} 0.986} \\
{\color[HTML]{000000} DORN \cite{fu2018deep}} & \multicolumn{1}{c}{{\color[HTML]{000000} $\surd$}} & {\color[HTML]{000000} $\surd$} & {\color[HTML]{000000} $>$51.0M} & {\color[HTML]{000000} -} & {\color[HTML]{000000} \textbf{0.072}} & {\color[HTML]{000000} \textbf{0.307}} & {\color[HTML]{000000} \textbf{2.727}} & {\color[HTML]{000000} \textbf{0.120}} & {\color[HTML]{000000} \textbf{0.932}} & {\color[HTML]{000000} \textbf{0.984}} & {\color[HTML]{000000} \textbf{0.994}} \\ \hline %\midrule
{\color[HTML]{000000} GeoNet-VGG \cite{yin2018geonet}} & {\color[HTML]{000000} } & {\color[HTML]{000000} -} & {\color[HTML]{000000} -} & {\color[HTML]{000000} -} & {\color[HTML]{000000} 0.164} & {\color[HTML]{000000} 1.303} & {\color[HTML]{000000} 6.090} & {\color[HTML]{000000} 0.247} & {\color[HTML]{000000} 0.765} & {\color[HTML]{000000} 0.919} & {\color[HTML]{000000} 0.968} \\
{\color[HTML]{000000} GeoNet-Resnet \cite{yin2018geonet}} & {\color[HTML]{000000} } & {\color[HTML]{000000} -} & {\color[HTML]{000000} 229.3M} & {\color[HTML]{000000} -} & {\color[HTML]{000000} 0.155} & {\color[HTML]{000000} 1.296} & {\color[HTML]{000000} 5.857} & {\color[HTML]{000000} 0.233} & {\color[HTML]{000000} 0.793} & {\color[HTML]{000000} 0.931} & {\color[HTML]{000000} 0.973} \\
{\color[HTML]{000000} DDVO \cite{wang2018learning}} & {\color[HTML]{000000} } & {\color[HTML]{000000} $\surd$} & {\color[HTML]{000000} -} & {\color[HTML]{000000} -} & {\color[HTML]{000000} 0.151} & {\color[HTML]{000000} 1.257} & {\color[HTML]{000000} 5.583} & {\color[HTML]{000000} 0.228} & {\color[HTML]{000000} 0.810} & {\color[HTML]{000000} 0.936} & {\color[HTML]{000000} 0.974} \\
{\color[HTML]{000000} SC-SfMLearner \cite{bian2019unsupervised}} & {\color[HTML]{000000} } & {\color[HTML]{000000} $\surd$} & {\color[HTML]{000000} 59.4M} & {\color[HTML]{000000} -} & {\color[HTML]{000000} 0.149} & {\color[HTML]{000000} 1.137} & {\color[HTML]{000000} 5.771} & {\color[HTML]{000000} 0.230} & {\color[HTML]{000000} 0.799} & {\color[HTML]{000000} 0.932} & {\color[HTML]{000000} 0.973} \\
{\color[HTML]{000000} Struct2depth \cite{casser2019depth}} & {\color[HTML]{000000} } & {\color[HTML]{000000} -} &{\color[HTML]{000000} 67.0M} &  {\color[HTML]{000000} -} & {\color[HTML]{000000} \underline{0.141}} & {\color[HTML]{000000} 1.026} & {\color[HTML]{000000} 5.291} & {\color[HTML]{000000} 0.215} & {\color[HTML]{000000} 0.816} & {\color[HTML]{000000} 0.945} & {\color[HTML]{000000} 0.979} \\
{\color[HTML]{000000} Jia et al. \cite{mypaper2021}}& {\color[HTML]{000000} } & {\color[HTML]{000000} $\surd$} &{\color[HTML]{000000} 57.6M} & {\color[HTML]{000000} 3.5G} & {\color[HTML]{000000} 0.144} & {\color[HTML]{000000} \textbf{0.966}} & {\color[HTML]{000000} 5.078} & {\color[HTML]{000000} 0.208} & {\color[HTML]{000000} 0.815} & {\color[HTML]{000000} 0.945} & {\color[HTML]{000000} \textbf{0.981}} \\
{\color[HTML]{000000} SGDepth \cite{klingner2020self}} & {\color[HTML]{000000} } & {\color[HTML]{000000} $\surd$} &{\color[HTML]{000000} -} &  {\color[HTML]{000000} -} & {\color[HTML]{000000} \textbf{0.128}} & {\color[HTML]{000000} 1.003} & {\color[HTML]{000000} 5.085} & {\color[HTML]{000000} 0.206} & {\color[HTML]{000000} \underline{0.853}} & {\color[HTML]{000000} 0.951} & {\color[HTML]{000000} 0.978} \\
{\color[HTML]{000000} Monodepth2 \cite{godard2019digging}} & {\color[HTML]{000000} } & {\color[HTML]{000000} $\surd$} &{\color[HTML]{000000} 59.4M} &  {\color[HTML]{000000} 3.5G} & {\color[HTML]{000000} \textbf{0.128}} & {\color[HTML]{000000} 1.087} & {\color[HTML]{000000} 5.171} & {\color[HTML]{000000} \underline{0.204}} & {\color[HTML]{000000} \textbf{0.855}} & {\color[HTML]{000000} 0.953} & {\color[HTML]{000000} 0.978} \\
{\color[HTML]{000000} DLNet-CC \cite{mypaper2}}& {\color[HTML]{000000} } &{\color[HTML]{000000} $\surd$} & {\color[HTML]{000000} 59.4M} & {\color[HTML]{000000} -} & {\color[HTML]{000000} \textbf{0.128}} & {\color[HTML]{000000} 0.990} & {\color[HTML]{000000} \underline{5.064}} & {\color[HTML]{000000} \textbf{0.202}} & {\color[HTML]{000000} 0.851} & {\color[HTML]{000000} \textbf{0.955}} & {\color[HTML]{000000} \underline{0.980}} \\ 
{\color[HTML]{000000} DLNet-CL \cite{mypaper2}} & {\color[HTML]{000000} } &{\color[HTML]{000000} $\surd$} & {\color[HTML]{000000} 63.1M} & {\color[HTML]{000000} -} & {\color[HTML]{000000} \textbf{0.128}} & {\color[HTML]{000000} \underline{0.979}} & {\color[HTML]{000000} \textbf{5.033}} & {\color[HTML]{000000} \textbf{0.202}} & {\color[HTML]{000000} 0.851} & {\color[HTML]{000000} \underline{0.954}} & {\color[HTML]{000000} \underline{0.980}} \\ \hline %\cline{1-11} %\cmidrule(l){2-12} 
{\color[HTML]{000000} SfMLearner \cite{zhou2017unsupervised}} & {\color[HTML]{000000} } & {\color[HTML]{000000} } & {\color[HTML]{000000} 126.0M} & {\color[HTML]{000000} -} & {\color[HTML]{000000} 0.208} & {\color[HTML]{000000} 1.768} & {\color[HTML]{000000} 6.856} & {\color[HTML]{000000} 0.283} & {\color[HTML]{000000} 0.678} & {\color[HTML]{000000} 0.885} & {\color[HTML]{000000} 0.957} \\
{\color[HTML]{000000} Yang et al. (J) \cite{yang2017unsupervised}} & {\color[HTML]{000000} } & {\color[HTML]{000000} } &{\color[HTML]{000000} 126.0M} &  {\color[HTML]{000000} -} & {\color[HTML]{000000} 0.182} & {\color[HTML]{000000} 1.481} & {\color[HTML]{000000} 6.501} & {\color[HTML]{000000} 0.267} & {\color[HTML]{000000} 0.725} & {\color[HTML]{000000} 0.906} & {\color[HTML]{000000} 0.963} \\
{\color[HTML]{000000} Monodepth2* \cite{godard2019digging}} & {\color[HTML]{000000} } & {\color[HTML]{000000} } &{\color[HTML]{000000} 59.4M} & {\color[HTML]{000000} 3.5G} & {\color[HTML]{000000} 0.144} & {\color[HTML]{000000} 1.059} & {\color[HTML]{000000} 5.289} & {\color[HTML]{000000} 0.217} & {\color[HTML]{000000} 0.824} & {\color[HTML]{000000} 0.945} & {\color[HTML]{000000} 0.976} \\
{\color[HTML]{000000} DLNet-LL \cite{mypaper2}} & {\color[HTML]{000000} } & {\color[HTML]{000000} } & {\color[HTML]{000000} 25.8M} & {\color[HTML]{000000} \textbf{1.3G}} & {\color[HTML]{000000} 0.141} & {\color[HTML]{000000} 1.060} & {\color[HTML]{000000} 5.247} & {\color[HTML]{000000} \textbf{0.215}} & {\color[HTML]{000000} 0.830} & {\color[HTML]{000000} 0.944} & {\color[HTML]{000000} 0.977} \\ 
{\color[HTML]{000000} \textbf{FSLNet-S}} & {\color[HTML]{000000} } & {\color[HTML]{000000} } & {\color[HTML]{000000} \textbf{5.5M}} & {\color[HTML]{000000} 3.2G} & {\color[HTML]{000000} \underline{0.135}} & {\color[HTML]{000000} \underline{0.973}} & {\color[HTML]{000000} \underline{5.084}} & {\color[HTML]{000000} \underline{0.208}} & {\color[HTML]{000000} \underline{0.840}} & {\color[HTML]{000000} \underline{0.948}} & {\color[HTML]{000000} \underline{0.978}} \\ 
{\color[HTML]{000000} \textbf{FSLNet-L}} & {\color[HTML]{000000} } & {\color[HTML]{000000} } & {\color[HTML]{000000} \underline{16.5M}} & {\color[HTML]{000000} 10.9G} & {\color[HTML]{000000} \textbf{0.128}} & {\color[HTML]{000000} \textbf{0.897}} & {\color[HTML]{000000} \textbf{4.905}} & {\color[HTML]{000000} \textbf{0.200}} & {\color[HTML]{000000} \textbf{0.852}} & {\color[HTML]{000000} \textbf{0.953}} & {\color[HTML]{000000} \textbf{0.980}} \\ 

\bottomrule
\end{tabular}
\label{kittiresults}
\end{table*}

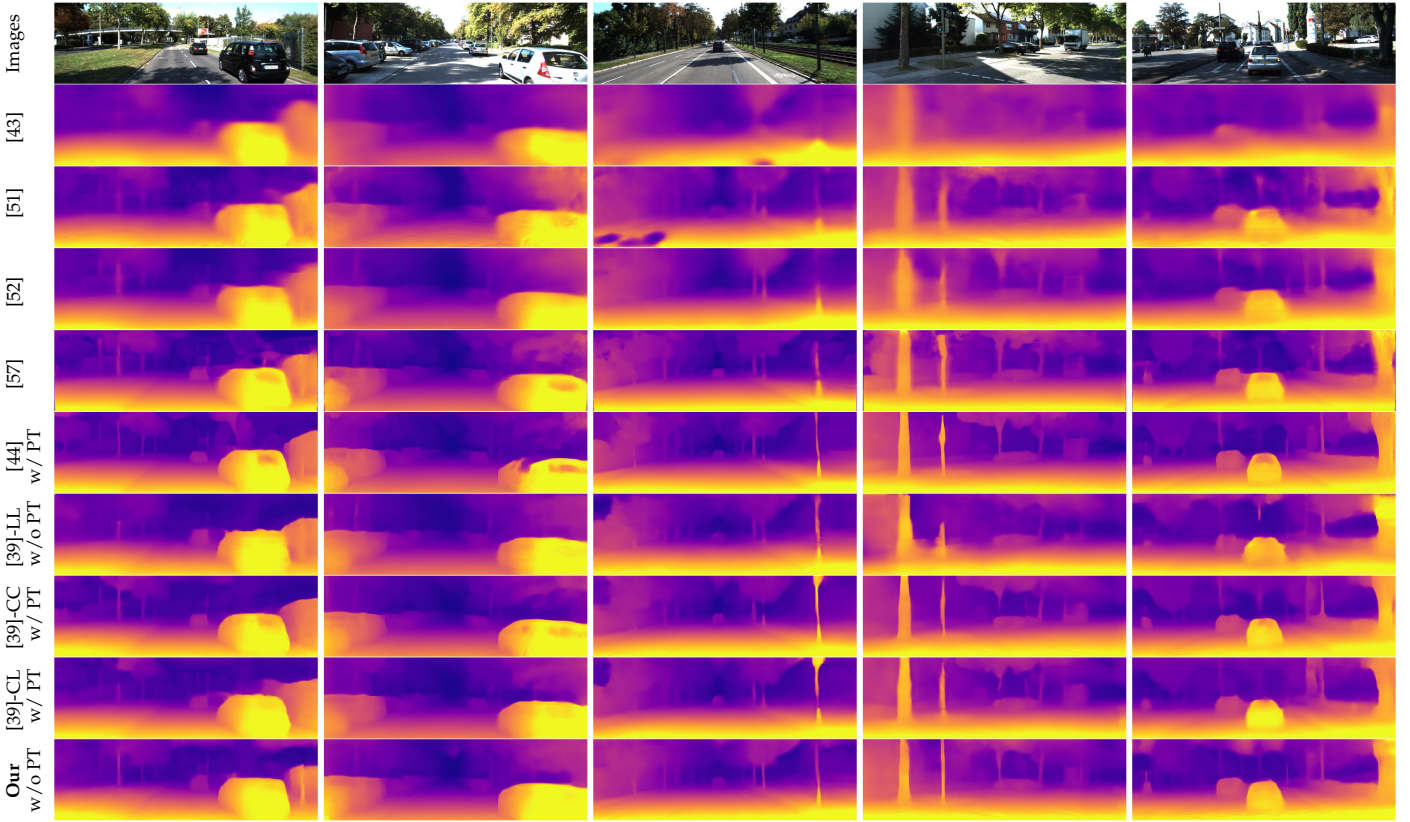
\begin{figure*}[htp]
    \centering
    \begin{overpic}[width=\textwidth]{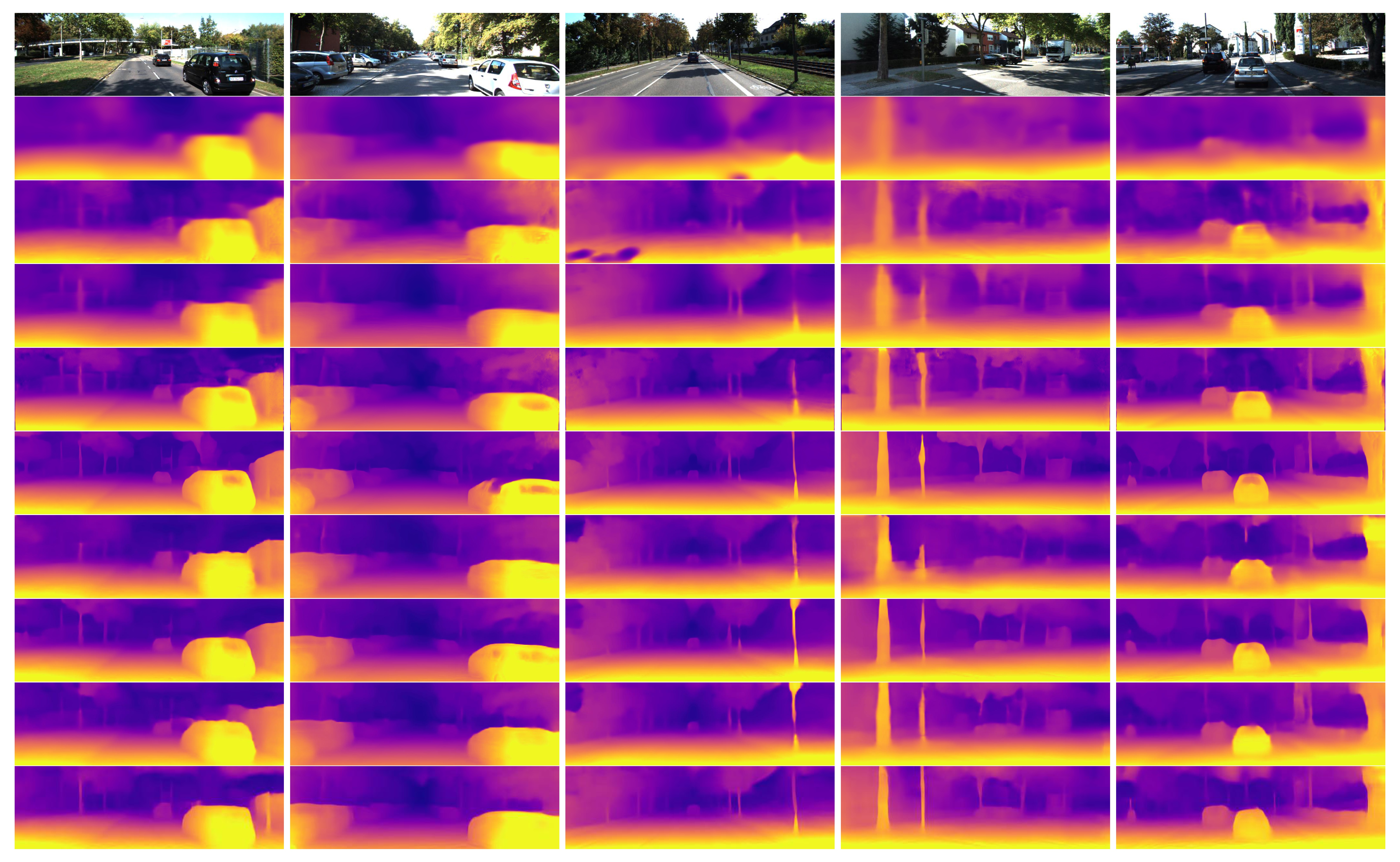}
    \put(-2.5,55.5){\scriptsize\rotatebox{90}{Images}}
    \put(-2.5,50){\scriptsize\rotatebox{90}{\cite{zhou2017unsupervised}}}
    \put(-2.5,44.5){\scriptsize\rotatebox{90}{\cite{yin2018geonet}}}
    \put(-2.5,38){\scriptsize\rotatebox{90}{\cite{wang2018learning}}}    
    \put(-2.5,32){\scriptsize\rotatebox{90}{\cite{godard2017unsupervised}}}
    \put(-2.5,26){\scriptsize\rotatebox{90}{\cite{godard2019digging}}} 
    \put(-1,25.5){\scriptsize\rotatebox{90}{w/ PT}} 
    \put(-2.5,19){\scriptsize\rotatebox{90}{\cite{mypaper2}-LL}}  
    \put(-1,19.5){\scriptsize\rotatebox{90}{w/o PT}} 
    \put(-2.5,13){\scriptsize\rotatebox{90}{\cite{mypaper2}-CC}} 
    \put(-1,14){\scriptsize\rotatebox{90}{w/ PT}} 
    \put(-2.5,7){\scriptsize\rotatebox{90}{\cite{mypaper2}-CL}} 
    \put(-1,8){\scriptsize\rotatebox{90}{w/ PT}}
    \put(-2.5,2.5){\scriptsize\rotatebox{90}{\textbf{Our}}} 
    \put(-1,1.5){\scriptsize\rotatebox{90}{w/o PT}}
    %\put(0,4){\scriptsize\rotatebox{90}{Our (CL)}}  
    %\put(1.5,3.1){\scriptsize\rotatebox{90}{w/ pretraining}} 
    \end{overpic}
    \caption{Qualitative results of depth estimation on the KITTI dataset \cite{geiger2013vision}. PT represents pretraining. The proposed model can correctly predict the challenging regions, such as cars' windows, fine fences, and fine trees, by extracting the global and local features simultaneously.}
    \label{qualitativeresultonkitti}
\end{figure*}

\begin{table*}[htb]
\caption{Depth estimation results on the Make3D dataset \cite{saxena2008make3d}. The best performances and our models are marked \textbf{bold}. The second-best performances in the second and third cells are \underline{underlined}.}
\small
\centering
\begin{tabular}{@{}lcccccc@{}}
\toprule
{\color[HTML]{000000} } & {\color[HTML]{000000} } & {\color[HTML]{000000} } & \multicolumn{4}{c}{{\color[HTML]{000000} Errors $\downarrow$}} \\ \cmidrule(l){4-7} 
\multirow{-2}{*}{{\color[HTML]{000000} Methods}} & \multirow{-2}{*}{{\color[HTML]{000000} GT?}} & \multirow{-2}{*}{{\color[HTML]{000000} PT?}} & {\color[HTML]{000000} AbsRel} & {\color[HTML]{000000} SqRel} & {\color[HTML]{000000} RMS} & {\color[HTML]{000000} RMSlog} \\ \midrule
{\color[HTML]{000000} Karsch et al. \cite{karsch2014depth}} & \multicolumn{1}{c}{{\color[HTML]{000000} $\surd$}} & {\color[HTML]{000000} } & {\color[HTML]{000000} 0.428} & {\color[HTML]{000000} 5.079} & {\color[HTML]{000000} 8.389} & {\color[HTML]{000000} 0.149} \\
{\color[HTML]{000000} Liu et al. \cite{liu2014discrete}} & \multicolumn{1}{c}{{\color[HTML]{000000} $\surd$}} & {\color[HTML]{000000} } & {\color[HTML]{000000} 0.475} & {\color[HTML]{000000} 6.562} & {\color[HTML]{000000} 10.05} & {\color[HTML]{000000} 0.165} \\
{\color[HTML]{000000} Laina et al. \cite{laina2016deeper}} & \multicolumn{1}{c}{{\color[HTML]{000000} $\surd$}} & {\color[HTML]{000000} $\surd$} & {\color[HTML]{000000} \textbf{0.204}} & {\color[HTML]{000000} \textbf{1.840}} & {\color[HTML]{000000} \textbf{5.683}} & {\color[HTML]{000000} \textbf{0.084}} \\
\midrule
{\color[HTML]{000000} DDVO \cite{wang2018learning}} & {\color[HTML]{000000} } & {\color[HTML]{000000} $\surd$} & {\color[HTML]{000000} 0.387} & {\color[HTML]{000000} 4.720} & {\color[HTML]{000000} 8.090} & {\color[HTML]{000000} 0.204} \\
{\color[HTML]{000000} Monodepth \cite{godard2017unsupervised}} & {\color[HTML]{000000} } & {\color[HTML]{000000} $\surd$} & {\color[HTML]{000000} 0.544} & {\color[HTML]{000000} 10.94} & {\color[HTML]{000000} 11.760} & {\color[HTML]{000000} \underline{0.193}} \\
{\color[HTML]{000000} Monodepth2 \cite{godard2019digging}} & {\color[HTML]{000000} } & {\color[HTML]{000000} $\surd$} & {\color[HTML]{000000} 0.322} & {\color[HTML]{000000} 3.589} & {\color[HTML]{000000} 7.417} & {\color[HTML]{000000} \textbf{0.163}} \\
{\color[HTML]{000000} Jia et al. \cite{mypaper2021}}& {\color[HTML]{000000} } & {\color[HTML]{000000} $\surd$} & {\color[HTML]{000000} 0.301} & {\color[HTML]{000000} 3.143} & {\color[HTML]{000000} 6.972} & {\color[HTML]{000000} 0.351} \\
{\color[HTML]{000000}  DLNet-CL \cite{mypaper2}}& {\color[HTML]{000000} } & {\color[HTML]{000000} $\surd$ } & {\color[HTML]{000000} \underline{0.269}} & {\color[HTML]{000000} \underline{2.201}} & {\color[HTML]{000000} \underline{6.452}} & {\color[HTML]{000000} 0.325} \\
{\color[HTML]{000000}  DLNet-CC \cite{mypaper2}}& {\color[HTML]{000000} } & {\color[HTML]{000000} $\surd$ } & {\color[HTML]{000000} \textbf{0.267}} & {\color[HTML]{000000} \textbf{2.188}} & {\color[HTML]{000000} \textbf{6.406}} & {\color[HTML]{000000} 0.322} \\
\midrule
{\color[HTML]{000000} SfMLearner \cite{zhou2017unsupervised}} & {\color[HTML]{000000} } & {\color[HTML]{000000} } & {\color[HTML]{000000} 0.383} & {\color[HTML]{000000} 5.321} & {\color[HTML]{000000} 10.470} & {\color[HTML]{000000} 0.478} \\
{\color[HTML]{000000}  DLNet-LL \cite{mypaper2}}& {\color[HTML]{000000} } & {\color[HTML]{000000} } & {\color[HTML]{000000} 0.289} & {\color[HTML]{000000} 2.423} & {\color[HTML]{000000} 6.701} & {\color[HTML]{000000} 0.348} \\
{\color[HTML]{000000}  \textbf{FSLNet-S}}& {\color[HTML]{000000} } & {\color[HTML]{000000} } & {\color[HTML]{000000} 0.282} & {\color[HTML]{000000} 2.376} & {\color[HTML]{000000} 6.476} & {\color[HTML]{000000} \textbf{0.336}} \\
{\color[HTML]{000000}  \textbf{FSLNet-L}}& {\color[HTML]{000000} } & {\color[HTML]{000000} } & {\color[HTML]{000000} 0.284} & {\color[HTML]{000000} 2.385} & {\color[HTML]{000000} 6.452} & {\color[HTML]{000000} \underline{0.337}} \\
{\color[HTML]{000000}  \textbf{FSLNet-S (Entire image input)}}& {\color[HTML]{000000} } & {\color[HTML]{000000} } & {\color[HTML]{000000} \underline{0.264}} & {\color[HTML]{000000} \textbf{1.755}} & {\color[HTML]{000000} \textbf{5.909}} & {\color[HTML]{000000} 0.341} \\
{\color[HTML]{000000}  \textbf{FSLNet-L (Entire image input)}}& {\color[HTML]{000000} } & {\color[HTML]{000000} } & {\color[HTML]{000000} \textbf{0.260}} & {\color[HTML]{000000} \underline{1.787}} & {\color[HTML]{000000} \underline{5.952}} & {\color[HTML]{000000} 0.343} \\

\bottomrule
\end{tabular}
\label{make3dresults}
\end{table*}

\begin{table*}[htb]
\caption{Absolute trajectory error (ATE) on the KITTI odometry dataset \cite{geiger2013vision} following Zhan’s split \cite{zhan2018unsupervised}. PT indicates pretraining.}
\small
\centering
\begin{tabular}{@{}lcccc@{}}
\toprule
{\color[HTML]{000000} Methods} & {\color[HTML]{000000} Pretraining} & {\color[HTML]{000000} Model size} & {\color[HTML]{000000} Seq. 09} & {\color[HTML]{000000} Seq. 10} \\ \midrule
{\color[HTML]{000000} Mean Odometry} & {\color[HTML]{000000} -} & {\color[HTML]{000000} -} & {\color[HTML]{000000} 0.032 $\pm$ 0.026} & {\color[HTML]{000000} 0.028 $\pm$ 0.023} \\
{\color[HTML]{000000} ORB-SLAM (short)} & {\color[HTML]{000000} -} & {\color[HTML]{000000} -} &  {\color[HTML]{000000} 0.064 $\pm$ 0.141} & {\color[HTML]{000000} 0.064 $\pm$ 0.130} \\
{\color[HTML]{000000} ORB-SLAM (full)} & {\color[HTML]{000000} -} &{\color[HTML]{000000} -} &  {\color[HTML]{000000} 0.014 $\pm$ 0.008} & {\color[HTML]{000000} 0.012 $\pm$ 0.011} \\
{\color[HTML]{000000} Vid2Depth \cite{mahjourian2018unsupervised}} & {\color[HTML]{000000} $\surd$} & {\color[HTML]{000000} 137.0M} &   {\color[HTML]{000000} 0.013 $\pm$ 0.010} & {\color[HTML]{000000} 0.012 $\pm$ 0.011} \\
{\color[HTML]{000000} Monodepth2 \cite{godard2019digging}} & {\color[HTML]{000000} $\surd$} &  {\color[HTML]{000000} 50.1M} & {\color[HTML]{000000} 0.017 $\pm$ 0.008} & {\color[HTML]{000000} 0.015 $\pm$ 0.010} \\
{\color[HTML]{000000} GeoNet \cite{yin2018geonet}} & {\color[HTML]{000000} -} & {\color[HTML]{000000} 229.3M} &  {\color[HTML]{000000} \underline{0.012 $\pm$ 0.007}} & {\color[HTML]{000000} \underline{0.012 $\pm$ 0.009}} \\
{\color[HTML]{000000} DLNet-CL \cite{mypaper2}} & {\color[HTML]{000000} $\surd$} &  {\color[HTML]{000000} 55.1M} & {\color[HTML]{000000} 0.017 $\pm$ 0.007} & {\color[HTML]{000000} 0.014 $\pm$ 0.007} \\
{\color[HTML]{000000} DLNet-CC \cite{mypaper2}} & {\color[HTML]{000000} $\surd$} &  {\color[HTML]{000000} 50.1} & {\color[HTML]{000000} 0.016 $\pm$ 0.007} & {\color[HTML]{000000} 0.014 $\pm$ 0.007} \\ 
{\color[HTML]{000000} Struct2depth \cite{casser2019depth}} & {\color[HTML]{000000} -} &  {\color[HTML]{000000} $>$50.1M} & {\color[HTML]{000000} \textbf{0.011 $\pm$ 0.006}} & {\color[HTML]{000000} \textbf{0.011 $\pm$ 0.010}} \\ \midrule
{\color[HTML]{000000} SfMLearner \cite{zhou2017unsupervised}} & {\color[HTML]{000000} } &  {\color[HTML]{000000} 137.0M} &  {\color[HTML]{000000} 0.021 $\pm$ 0.017} & {\color[HTML]{000000} 0.020 $\pm$ 0.015} \\
{\color[HTML]{000000} DLNet-LL \cite{mypaper2}} & {\color[HTML]{000000} } &  {\color[HTML]{000000} 17.9M} & {\color[HTML]{000000} \underline{0.018 $\pm$ 0.006}} & {\color[HTML]{000000} \underline{0.015 $\pm$ 0.007}} \\
{\color[HTML]{000000}  \textbf{FSLNet-S (Our)}} & {\color[HTML]{000000} } &  {\color[HTML]{000000} \textbf{3.9M}} & {\color[HTML]{000000} \textbf{0.014 $\pm$ 0.007}} & {\color[HTML]{000000} \textbf{0.011 $\pm$ 0.007}} \\
{\color[HTML]{000000}  \textbf{FSLNet-L (Our)}} & {\color[HTML]{000000} } &  {\color[HTML]{000000} \underline{11.9M}} & {\color[HTML]{000000} \textbf{0.014 $\pm$ 0.007}} & {\color[HTML]{000000} \textbf{0.011 $\pm$ 0.007}} \\

\bottomrule
\end{tabular}
\label{visualodometry}
\end{table*}

\subsubsection{Loss functions}
During training, we follow the loss functions presented in \cite{mypaper2} except for the multi-scale prediction. The final loss, $\mathcal{L}$, consists of three parts: reconstruction loss $\mathcal{L}_{recons}$, 3D geometry smoothness (3DGS) loss $\mathcal{L}_{3DGS}$, and the proposed self-contrast loss $\mathcal{L}_{scontr}$; it can be described as follows (Eq. \ref{totalloss}):
\begin{equation}
\label{totalloss}
    \mathcal{L} = \mathcal{L}_{recons} + \alpha\mathcal{L}_{3DGS} + \beta\mathcal{L}_{scontr},
\end{equation}
wherein $\alpha$ and $\beta$ are weighting factors, being set to $10^{-3}$ in this paper. 

\textbf{Reconstruction loss.} Let $I_{t}$ and $I_{i}, i=-k,...,-1,1,...,k$ be target image and reference images. The reconstruction loss is to describe the difference between the reconstructed image and the target image. Godard et al. \cite{godard2019digging} proposed minimum reprojection and auto-masking to improve the visual inconsistency issue existing in the target and reference images and the relatively stationary problem between the objects and camera. We integrate these into reconstruction loss, which can be described as Eq. \ref{reconos}.
\begin{equation}
\label{reconos}
    \mathcal{L}_{recons} = \frac{1}{N_{b}}\sum_{b}\frac{1}{N_{p}}\sum_{p}L_{recons},
\end{equation}
where $N_{b}$, $N_{p}$, $b$, and $p$ represent the batch size, the number of the pixels, the traversing of each sample, and the traversing of each pixel; $L_{recons}$ is stated as Eq. \ref{reconos1},
\begin{equation}
\label{reconos1}
    L_{recons} = min(\mu_{-k}L_{-k},...,\mu_{-1}L_{-1}, \mu_{1}L_{1}, ..., \mu_{k}L_{k}).
\end{equation}
In Eq. \ref{reconos1}, $\mu_{i}, i=-k,...,-1, 1,...,k$ is the binary mask proposed by Godard et al. \cite{godard2019digging}, which can be defined as Eq. \ref{automasking}:
\begin{equation}
\label{automasking}
    \mu_{i} = [L_{i}<L_{io}],
\end{equation}
wherein $[\cdot]$ represents the Iverson bracket; $L_{i}$ is the difference between the reconstructed image and the target image, which is defined as follows Eq.\ref{recons2}; $L_{io}$ represents the difference between the reference images and the target image, which is also computed by Eq. \ref{recons2}. 
\begin{equation}
\label{recons2}
    L_{i} = 0.15L_{p}^{i}+0.85L_{ssim}^{i}.
\end{equation}
In Eq. \ref{recons2}, $L_{p}$ and $L_{ssim}$ represent the photometric loss and structure similarity (SSIM) loss, which can be stated as Eq. \ref{recons3}:
\begin{equation}
\label{recons3}
    \begin{aligned}
    L_{p}^{i} &= ||I_{i \to t} - I_{t}||_{1} \\
    L_{ssim}^{i} &= \frac{1-SSIM(I_{i \to t}, I_{t})}{2}
    \end{aligned},
\end{equation}
wherein and $SSIM(\cdot)$ represent the SSIM loss; $I_{i \to t}$ represents the reconstructed target images from the reference images according to Eq. \ref{viewsynthesis}.
\begin{equation}
\label{viewsynthesis}
    p_{r} \sim KE_{t \to i}D_{t}(p_{t})K^{-1}p_{t},
\end{equation}
wherein $p_{r}$, $K$, $E_{t \to i}$, $D_{t}(p_{t})$, $K^{-1}$, and $p_{t}$ represent the coordinate of a pixel in the reference image, the camera intrinsic matrix ($\mathbb{R}^{3 \times 3}$), the transform matrix between the target image and reference image, the depth corresponding to $p_{t}$, the inverse matrix of $K$, and the coordinate of a pixel in the target image, respectively.

\textbf{3D geometry smoothness loss.} 3D geometry smoothness is proposed by Jia et al. \cite{mypaper2}, aiming at imposing high-order smoothness constraints on the predicted depth map. We first convert the predicted depth map to 3D point clouds in terms of Eq. \ref{recoverpoints}
\begin{equation}
\label{recoverpoints}
    P \sim DK^{-1}p,
\end{equation}
wherein $P$, $D$, $K$, and $p$ represent the point clouds, the predicted depth map, the camera intrinsic matrix, and the image coordinates, respectively.

Then, we compute the surface normals for each point cloud. Given a target point $P_{t}$ and $8$ neighboring points $P_{i}, i=1,2,...,8$, the surface normal $\overrightarrow{n_{t}}$ at $P_{t}$ can be estimated by Eq. \ref{normal}:
\begin{equation}
\label{normal}
  \begin{aligned}
  \overrightarrow{n_{t}} &= \frac{1}{8} \sum_{i=1}^{8} \overrightarrow{n_i}, \\
  where \quad \overrightarrow{n_i} &= \overrightarrow{P_{t}P_{i}} \otimes \overrightarrow{P_{t}P_{j}}, \\
  j &= \begin{cases}
  i + 1 & i + 1 \leq 8 \\
  1 & i + 1 > 8
  \end{cases},
   \end{aligned}
\end{equation}
where $\otimes$ represents the cross-product operation. Moreover, the distance between normals is measured by sine distance following \cite{mypaper2}, which can be defined as Eq. \ref{sinedistance}:
\begin{equation}
\label{sinedistance}
\S(\overrightarrow{n_{1}}, \overrightarrow{n_{2}}) = 1 - ( \frac{\overrightarrow{n_{1}} \cdot \overrightarrow{n_{2}}}{\|\overrightarrow{n_{1}}\|_{2}\|\overrightarrow{n_{2}}\|_{2}})^{2},
\end{equation}
where $\S$ is the sine distance operator. Thus, the 3DGS loss can be described as Eq. \ref{geometrysmoothness} \cite{mypaper2}:
\begin{equation}
\label{geometrysmoothness}
\begin{aligned}
\mathcal{L}_{3DGS} = \frac{1}{N_b}\sum_{b}\frac{1}{N_p}&\sum_{p} \begin{matrix}  |e^{-\partial_{x}I}|\S_{x}A + |e^{-\partial_{y}I}|\S_{y}A \\
\end{matrix} \\
& + \begin{matrix} |e^{-\partial_{x}I}|\partial_{x}D^{-1} + |e^{-\partial_{y}I}|\partial_{y}D^{-1} \\
\end{matrix},
\end{aligned}
\end{equation}
where $\partial$, $A$ $D^{-1}$, and $I$ represent the gradient operator, the estimated surface normal matrix, the predicted disparity map, and the color image, respectively.

\textbf{Self-contrast loss.} Previous works adopt multi-scale prediction \cite{godard2019digging, mypaper2} to overcome the gradient locality issues caused by low-texture regions of the image. However, the multi-scale prediction will make the network more complex and raise the computational and storage costs. Accordingly, we propose a very simple but effective loss item, self-contrast loss, to overcome the gradient locality issue. Because the loss functions are solely used during training; thus, the proposed self-contrast loss does not influence the inference phase.

Specifically, we realize that some low-texture regions, such as road and sky areas, are mistakenly predicted because of the gradient locality issue; however, careful observations indicate that the problematic regions gradually appear along with the training process and these regions are small. Thus, we can perform morphological operations to the predicted disparity map, by which we can remove the incorrect estimates, making the predictions consistent. Thereafter, we can minimize the difference between the original predictions and the process predictions to overcome the gradient locality issue. Mathematically, we can define the proposed self-contrast loss $\mathcal{L}_{scontr}$ as Eq. \ref{selfcontrast}:
\begin{equation}
\label{selfcontrast}
    \begin{aligned}
    \mathcal{L}_{scontr} &= \frac{1}{N_b}\sum_{b}\frac{1}{N_{ep}}\sum_{ep} [D_{diff} > \epsilon] D_{diff} \\
    D_{diff} &= ||D - opening(D)||_{1} 
    \end{aligned},
\end{equation}
wherein $[\cdot]$, $N_{b}$, $N_{ep}$, $b$, and $ep$ represent the Iverson bracket, the batch size, the number of the effective pixels determined by $[D_{diff} > \epsilon]$, the traversing of each sample, and the traversing of each effective pixel determined by $[D_{diff} > \epsilon]$; $opening$ represents the opening operation with a $1$-valued kernel and the kernel size being $31 \times 31$; $\epsilon$ is the threshold and we adopt $0.3 \times 
(max(D) - min(D))$ in this paper.

\subsubsection{Datasets}

For depth estimation, we use the KITTI dataset \cite{geiger2013vision}, a large-scale and comprehensive computer vision task benchmarks, to train the networks. Following \cite{zhou2017unsupervised}, we use $40,109$ and $4,431$ $3$-frame sequences for training and validation, respectively. Following \cite{eigen2014depth}, we take $697$ images from the KITTI dataset \cite{geiger2013vision} for testing. 

Additionally, we directly apply the model trained on the KITTI dataset \cite{geiger2013vision} to the Make3D dataset \cite{saxena2008make3d} to validate the generalization capability of the proposed model. Here, Make3D, having 134 testing images, is an outdoor dataset as well, but has different scenes, color distributions, and perspectives; thus, it is challenging to perform prediction. Following \cite{godard2019digging}, we evaluate Make3D's images on a center crop at a $2 \times 1$ ratio.

For ego-motion estimation, we use the KITTI odometry dataset \cite{geiger2013vision} to evaluate the proposed model. Following \cite{zhan2018unsupervised}, the sequences $00$-$08$ and $09$-$10$ in the KITTI odometry dataset \cite{geiger2013vision} are used for training and testing, respectively. We use $3$-frame snippets as the input of the pose estimation network. Note that depth estimation and pose estimation are separately trained.

\subsubsection{Implementation details}

\textbf{Prediction blocks.} We simply adopt three convolutional layers as the prediction block for both depth and ego-motion estimation. Specifically, two single-layer CNN blocks as presented in Section $4.2$ followed by a single convolutional layer without activation layer are adopted for depth estimation; the input-output channel pairs for these three layers are $(c, \frac{c}{2})$, $(\frac{c}{2}, \frac{c}{4})$, and $(\frac{c}{4}, 1)$, respectively. 

For the ego-motion estimation network, we adopt the encoder part of the proposed FSLNet as the backbone. Then, a similar prediction block, with the input-output channel pairs $(8c, 4c$, $(4c, 2c)$, and $2c, 12)$, is utilized to predict the pose vectors.

\textbf{Data augmentation.} Following \cite{godard2019digging}, Random cropping, scaling, and horizontal flips are used to augment the input images. Additionally, a series of color augmentations, i.e., random brightness, contrast, saturation, and hue jitter with respective ranges of $\pm0.2$, $\pm0.2$, $\pm0.2$, and $\pm0.1$, are adopted with a $50$ percent chance. Note that these color augmentations are solely applied to the images fed to the network, not to those applied to compute the loss.

\textbf{Hyperparameters.} The networks are trained using the Adam optimizer \cite{kingma2014adam} with the initial learning rate $10^{-4}$; the learning rate decreases by a factor of $10$ every $15$ epochs. The other parameters of the Adam optimizer are set to the default values. The epoch, batch size, length of the sequence, and image resolution are set to $20$, $12$, $3$, $416 \times 128$, unless specified. During the training phase, we randomly initialize all parameters, without any pretraining or special initialization scheme.

\subsection{Semantic segmentation}
\subsubsection{Task description}
Semantic segmentation is a high-level computer vision task compared with depth estimation, which aims at predicting semantic labels for each pixel in the image. Tremendous work has been done in semantic segmentation and achieved promising results. In this work, the focus is not to pursue the state-of-the-art performance in semantic segmentation but to demonstrate the effectiveness of the proposed model, FSLNet. Therefore, we do not compare the results with other state-of-the-art methods. Instead, we choose the PyTorch implementation of DeepLabv3 \cite{chen2017rethinking, pytorchdeeplibv3} as the baseline model; then we gradually substitute the encoder and decoder in DeepLabv3 with the proposed model to observe the performance changes.

To formulate semantic segmentation, let $I \in \mathbb{R}^{H \times W \times 3}$ and $S \in \mathbb{R}^{H \times W \times C}$ represent the input image and the predicted labels, wherein $H$, $W$, and $C$ represent the height of the input image, the width of the input image, and the number of classes, respectively. Thus, semantic segmentation can be described as: $G_{S}: I \in \mathbb{R}^{H \times W \times 3} \to S \in \mathbb{R}^{H \times W \times C}$. We use the cross-entropy loss to train the network.

\subsubsection{Dataset}

Cityscapes \cite{cordts2016cityscapes}, a large-scale semantic segmentation dataset, is used to evaluate the proposed FSLNet. Cityscapes contains high-quality pixel-wise annotations of $5,000$ images, in which $2,975$, $500$, and $1,525$ images are used for training, validation, and testing, respectively. Moreover, $20,000$ coarsely annotated images are also provided. In this paper, we train the model solely using the pixel-wise annotated $2,975$ images; then we evaluate the performance on the validation dataset following \cite{cordts2016cityscapes}.

\subsubsection{Implementations details}

\textbf{Prediction block.} Semantic segmentation is a multiple classification task, i.e., the number of output channel is equal to the number of classes, instead of being $1$ as shown in depth estimation. Thus, we simply adjust the input-output channel pairs of the Prediction block used for depth estimation; specifically, we change the input-output channel pairs $(c, \frac{c}{2})$, $(\frac{c}{2}, \frac{c}{4})$, and $(\frac{c}{4}, 1)$ to $(c, C$, $(C, C)$, and $(C, C)$. Other configurations are set as identical with the Prediction block of depth estimation.

\textbf{Data augmentation.} Following \cite{chen2017rethinking}, we randomly scale the input image with the scaling factor in the range of $0.5$ to $2$.  Moreover, we do a left-right flip to the input image with a $50$ percent chance.

\textbf{Hyperparameters.} The networks are trained using the Adam optimizer \cite{kingma2014adam} with the initial learning rate $10^{-4}$. The other parameters of the Adam optimizer are set to the default values. The batch size and the image resolution are set to $3$ and $1024 \times 512$ following \cite{pytorchdeeplibv3}. When training the model "FSLNet-L encoder+ASPP (Atrous Spatial Pyramid Pooling, the decoder proposed in \cite{chen2017rethinking})", we initially set the epoch to $1,000$ following \cite{pytorchdeeplibv3}. Once we observe the performance being better than those presented in \cite{pytorchdeeplibv3}, the training ends. While training the proposed FSLNet-S and FSLNet-L, we set the epoch to be $120$. Note that we treat the validation dataset as the test dataset, i.e., we only train and test these models once and do not adjust any parameters during training.

\begin{figure}[htp]
    \centering
    \begin{overpic}[scale=0.58]{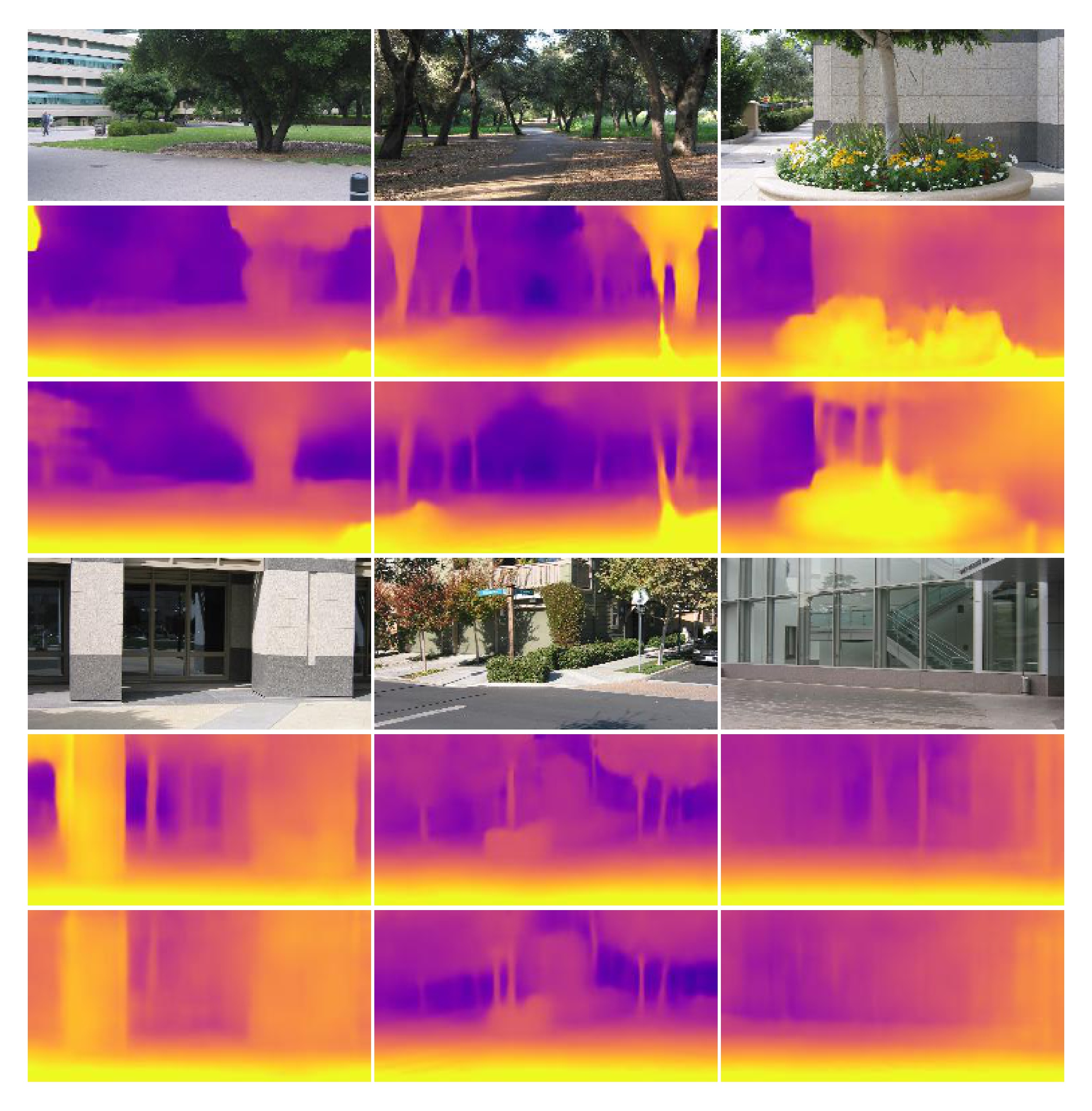}
    \put(-1.5,85){\scriptsize\rotatebox{90}{Images}}
    \put(-1.5,66){\scriptsize\rotatebox{90}{DLNet \cite{mypaper2}}}
    \put(-1.5,55){\scriptsize\rotatebox{90}{\textbf{Our}}}
    \put(-1.5,37){\scriptsize\rotatebox{90}{Images}}
    \put(-1.5,19){\scriptsize\rotatebox{90}{DLNet \cite{mypaper2}}}
    \put(-1.5,7){\scriptsize\rotatebox{90}{\textbf{Our}}}
    %\put(0,92){\scriptsize\rotatebox{90}{Images}}
    %\put(0,80){\scriptsize\rotatebox{90}{SfMLearner %\cite{zhou2017unsupervised}}}
    %\put(12,12){\color[rgb]{1,0,0}\oval(12,12)}

    \end{overpic}
    \caption{Qualitative results of depth estimation on the Make3D dataset \cite{saxena2008make3d} (cropped images).}
    \label{qualitativeresultsonmake3d}
\end{figure}

\begin{figure}[htp]
    \centering
    \begin{overpic}[scale=1.1]{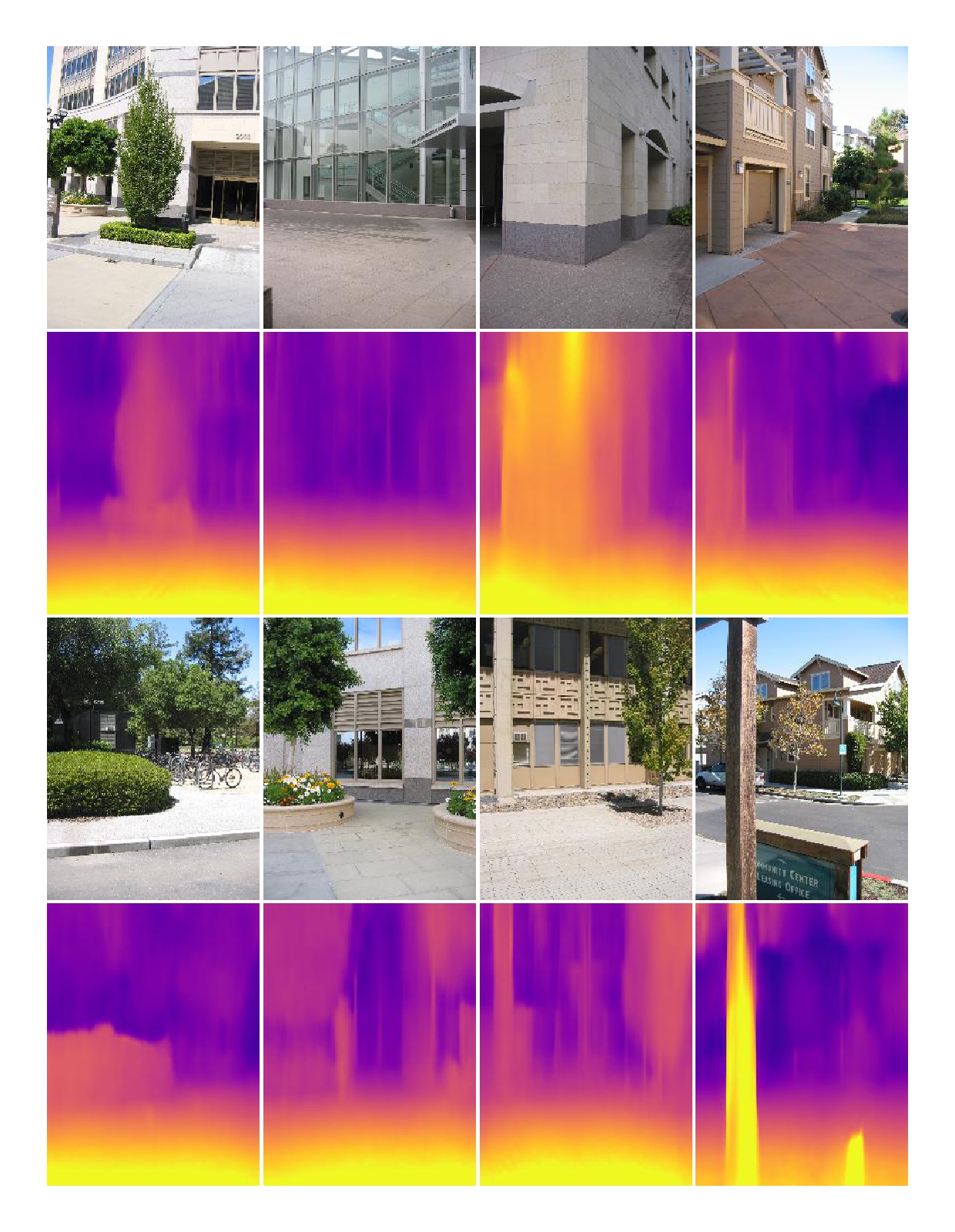}
    \put(-0.5,82){\scriptsize\rotatebox{90}{Images}}
    \put(-0.5,58){\scriptsize\rotatebox{90}{\textbf{Our}}}
    \put(-0.5,34){\scriptsize\rotatebox{90}{Images}}
    \put(-0.5,12){\scriptsize\rotatebox{90}{\textbf{Our}}}
    %\put(0,92){\scriptsize\rotatebox{90}{Images}}
    %\put(0,80){\scriptsize\rotatebox{90}{SfMLearner %\cite{zhou2017unsupervised}}}
    %\put(12,12){\color[rgb]{1,0,0}\oval(12,12)}

    \end{overpic}
    \caption{Qualitative results of depth estimation on the Make3D dataset \cite{saxena2008make3d} (entire images).}
    \label{qualitativeresultsonmake3dfull}
\end{figure}
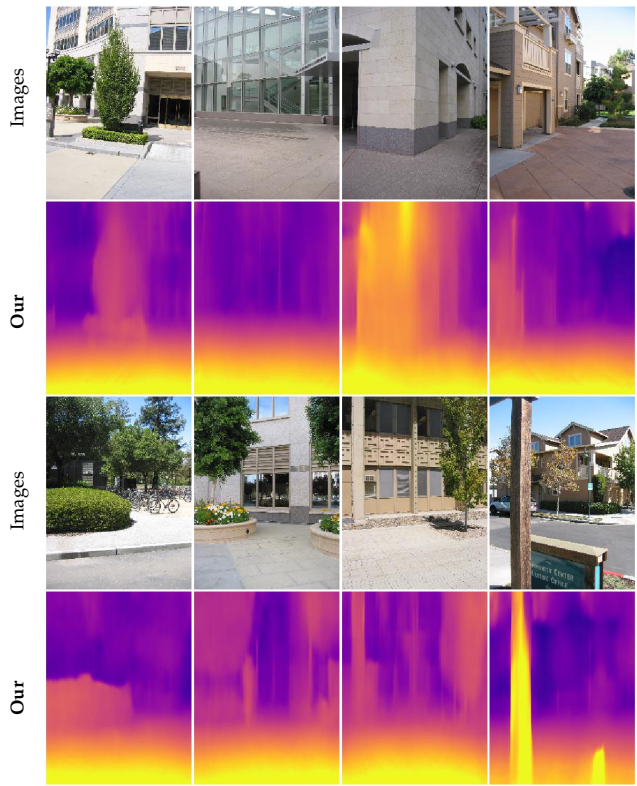

\begin{table*}[htb]
\caption{Quantitative results of semantic segmentation (Intersection over Union (IOU)) on the Cityscapes dataset \cite{cordts2016cityscapes}. The best performances in each row are marked \textbf{bold}. ASPP represents Atrous Spatial Pyramid Pooling proposed in \cite{chen2017rethinking}.}
\small
\centering
\begin{tabular}{@{}lcccc@{}}
\toprule
Items & Baseline \cite{pytorchdeeplibv3} & FSLNet-L encoder + ASPP & Full FSLNet-S & Full FSLNet-L\\ \midrule
road & 0.918 & 0.953 & 0.974 & \textbf{0.978}\\
sidewalk & 0.715 & 0.685 & 0.795 & \textbf{0.818}\\
building & 0.837 & 0.861 & 0.889 & \textbf{0.904}\\ 
wall & \textbf{0.413} & 0.280 & 0.329 & 0.387\\ 
fence & 0.397 & 0.480 & 0.487 & \textbf{0.530} \\ 
pole & 0.404 & 0.299 & 0.515 & \textbf{0.570}\\ 
traffic light & 0.411 & 0.480 & 0.474 & \textbf{0.590}\\ 
traffic sign & 0.577 & 0.554 & 0.613 & \textbf{0.698}\\ 
vegetation & 0.857 & 0.867 & 0.899 & \textbf{0.909}\\
terrain & 0.489 & 0.519 & 0.545 & \textbf{0.570}\\
sky & 0.850 & 0.889 & 0.938 &  \textbf{0.941}\\ 
person & 0.637 & 0.616 & 0.686 & \textbf{0.743}\\ 
rider & 0.456 & 0.409 & 0.372 & \textbf{0.464}\\ 
car & 0.897 & 0.847 & 0.911 & \textbf{0.929}\\ 
truck & 0.582 & 0.511 & 0.558 & \textbf{0.620}\\ 
bus & 0.616 & 0.638 & 0.596 & \textbf{0.671}\\ 
train & 0.310 & 0.584 & 0.395 & \textbf{0.437}\\ 
motorcycle & 0.322 & 0.332 & 0.254 & \textbf{0.377}\\ 
bicycle & 0.583 & 0.586 & 0.636 & \textbf{0.683}\\ \midrule
average & 0.593 & 0.599 & 0.625 & \textbf{0.675}\\
\# of epochs & 580 & 340 & 120 & 120 \\
model size & 58.5M & 14.6M & 5.6M & 16.5M\\
\bottomrule
\end{tabular}
\label{semanticsegmentationresults}
\end{table*}

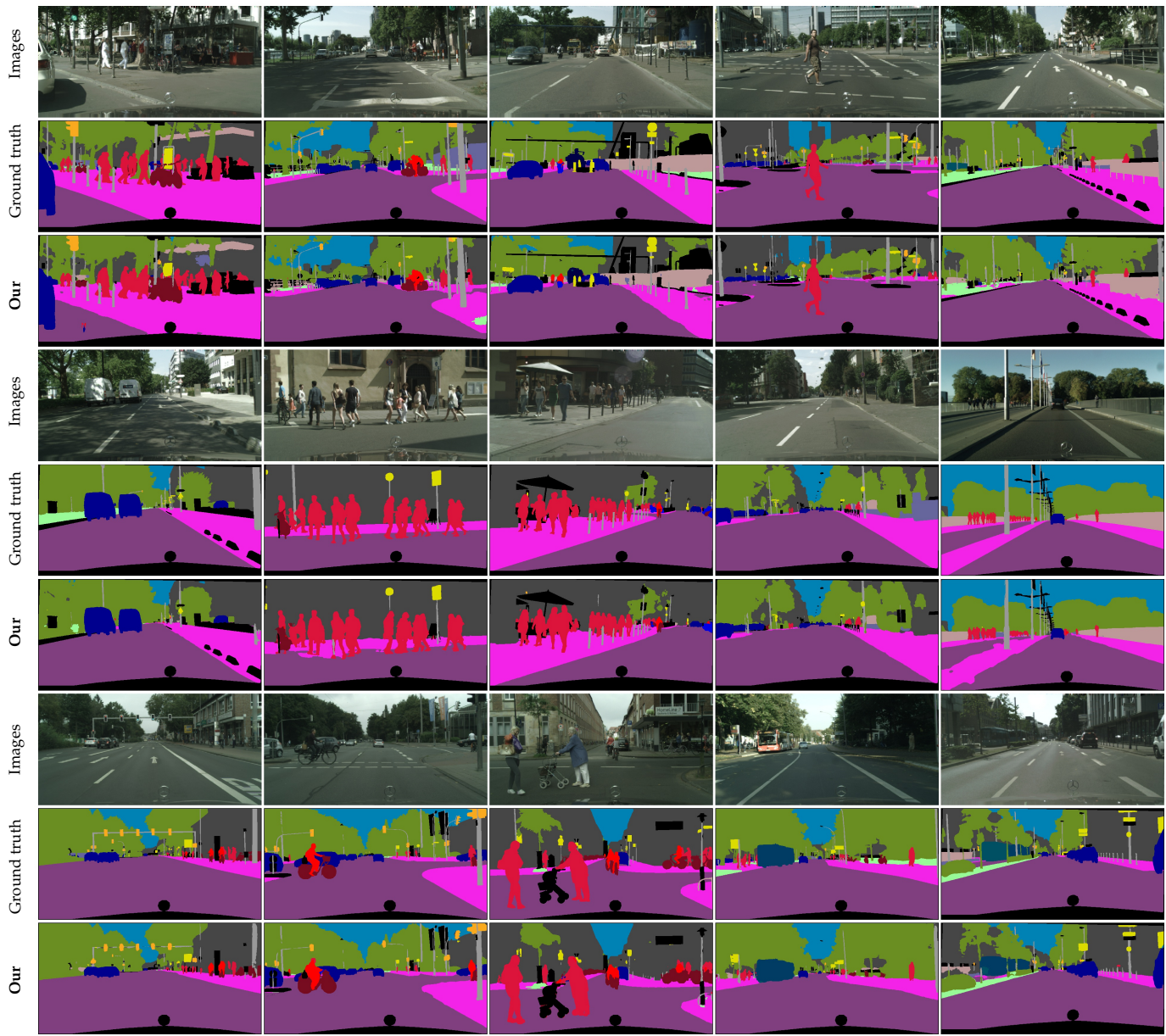
\begin{figure*}[htp]
    \centering
    \begin{overpic}[width=\textwidth]{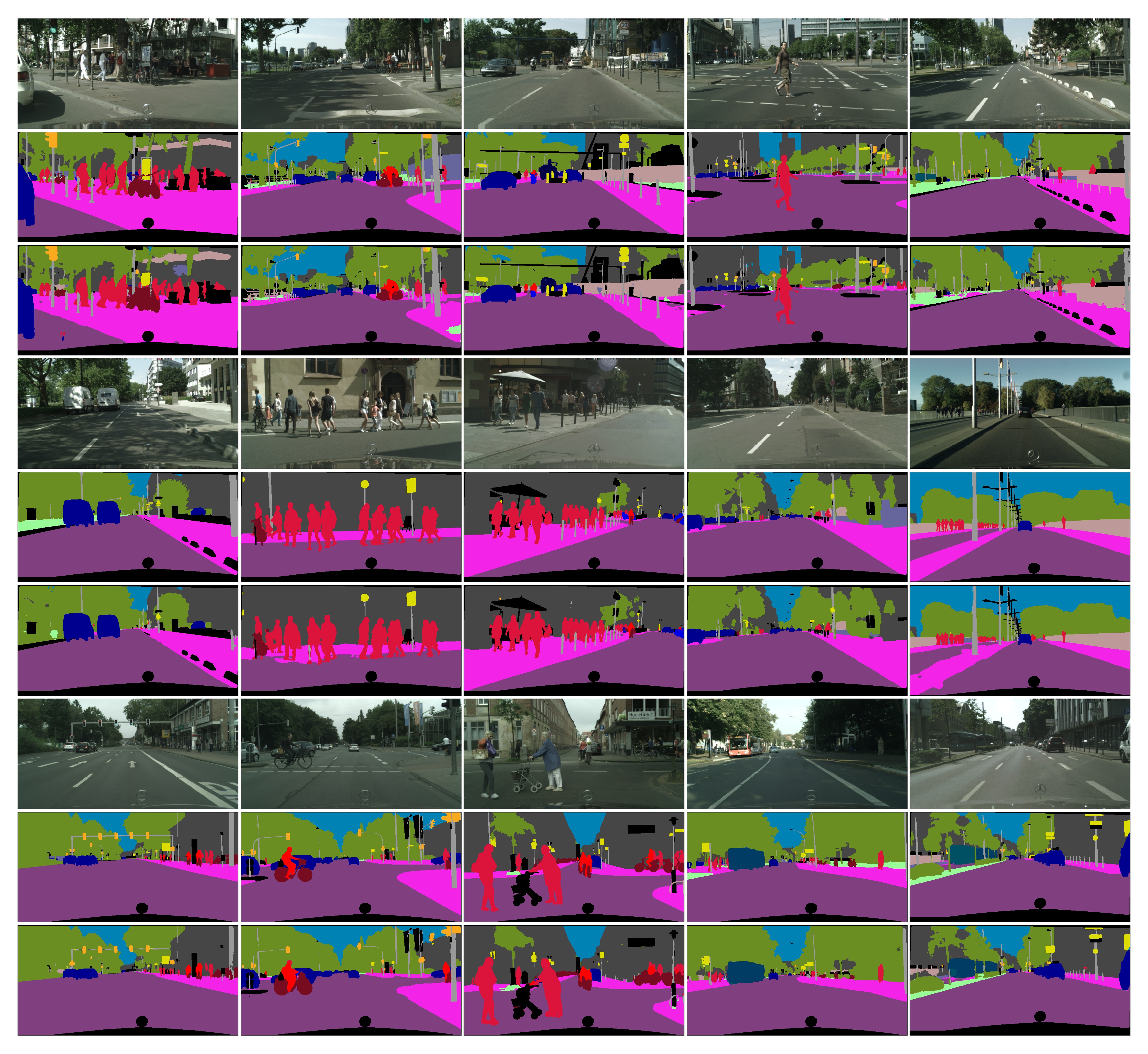}
    \put(-1,84){\scriptsize\rotatebox{90}{Images}}
    \put(-1,72){\scriptsize\rotatebox{90}{Ground truth}}
    \put(-1,64){\scriptsize\rotatebox{90}{\textbf{Our}}}
    \put(-1,54){\scriptsize\rotatebox{90}{Images}}
    \put(-1,42){\scriptsize\rotatebox{90}{Ground truth}}
    \put(-1,35){\scriptsize\rotatebox{90}{\textbf{Our}}}
    \put(-1,24){\scriptsize\rotatebox{90}{Images}}
    \put(-1,12){\scriptsize\rotatebox{90}{Ground truth}}
    \put(-1,5){\scriptsize\rotatebox{90}{\textbf{Our}}}    
    %\put(0,92){\scriptsize\rotatebox{90}{Images}}
    %\put(0,80){\scriptsize\rotatebox{90}{SfMLearner %\cite{zhou2017unsupervised}}}
    %\put(12,12){\color[rgb]{1,0,0}\oval(12,12)}

    \end{overpic}
    \caption{Qualitative results of semantic segmentation on the Cityscapes dataset \cite{cordts2016cityscapes}.}
    \label{qualitativeresultsoncity}
\end{figure*}

\begin{table*}[h]
\centering
\caption{Experiments on different activation layers. The best performances are marked \textbf{bold}.}
\begin{tabular}{lccccccc}
\toprule
\multirow{2}{*}{Activation layers} & \multicolumn{4}{c}{Errors $\downarrow$} & \multicolumn{3}{c}{Errors $\uparrow$} \\ \cline{2-8} 
 & AbsRel & SqRel & RMS & RMSlog & $<1.25$ & $<1.25^{2}$ & $<1.25^{3}$ \\ \hline
ELU & 0.152 & \textbf{1.057} & \textbf{5.245} & 0.220 & 0.810 & \textbf{0.938} & 0.975 \\
ReLU & 0.153 & \textbf{1.107} & 5.385 & \textbf{0.223} & 0.805 & 0.935 & 0.974 \\ 
\textbf{SiLU} & \textbf{0.150} & 1.076 & 5.292 & \textbf{0.219} & \textbf{0.812} & \textbf{0.938} & \textbf{0.976} \\ \bottomrule
\end{tabular}
\label{actlayers}
\end{table*}

\begin{table*}[h]
\centering
\caption{Experiments on different global-local ratios in FSLBlock (i.e., the ratios of the number of linear frequency learning layers and the number of convolutional layers). The best performances in different cells are marked \textbf{bold}.}
\begin{tabular}{lccccccc}
\toprule
\multirow{2}{*}{Global-local ratios} & \multicolumn{4}{c}{Errors $\downarrow$} & \multicolumn{3}{c}{Errors $\uparrow$} \\ \cline{2-8} 
 & AbsRel & SqRel & RMS & RMSlog & $<1.25$ & $<1.25^{2}$ & $<1.25^{3}$ \\ \hline
1:1 & 0.153 & 1.163 & 5.488 & 0.224 & 0.804 & 0.936 & 0.975 \\
2:2 & 0.151 & 1.100 & 5.351 & 0.221 & 0.808 & 0.938 & \textbf{0.976} \\ 
\textbf{3:3} & \textbf{0.150} & 1.076 & 5.292 & \textbf{0.219} & \textbf{0.812} & 0.938 & \textbf{0.976} \\ 
4:4 & 0.152 & \textbf{1.060} & \textbf{5.286} & 0.220 & 0.807 & \textbf{0.939} & \textbf{0.976} \\ 
5:5 & 0.155 & 1.119 & 5.318 & 0.222 & 0.804 & 0.935 & 0.974 \\ \midrule
3:1 & 0.153 & 1.124 & 5.399 & 0.223 & 0.805 & 0.936 & 0.975 \\
3:2 & 0.159 & 1.152 & 5.479 & 0.228 & 0.795 & 0.935 & 0.975 \\
3:3 & 0.150 & 1.076 & 5.292 & 0.219 & \textbf{0.812} & 0.938 & \textbf{0.976} \\
\textbf{3:4} & 0.150 & 1.063 & \textbf{5.265} & \textbf{0.218} & \textbf{0.812} & \textbf{0.940} & \textbf{0.976} \\
3:5 & \textbf{0.149} & \textbf{1.049} & 5.270 & 0.220 & 0.811 & 0.939 & 0.975 \\ \midrule
1:4 & 0.153 & 1.109 & 5.361 & 0.221 & 0.802 & 0.937 & \textbf{0.976} \\
\textbf{2:4} & \textbf{0.148} & \textbf{1.047} & \textbf{5.217} & \textbf{0.218} & \textbf{0.815} & \textbf{0.940} & \textbf{0.976} \\
3:4 & 0.150 & 1.063 & 5.265 & \textbf{0.218} & 0.812 & \textbf{0.940} & \textbf{0.976} \\
4:4 & 0.152 & 1.060 & 5.286 & 0.220 & 0.807 & 0.939 & \textbf{0.976} \\
5:4 & 0.155 & 1.122 & 5.360 & 0.222 & 0.804 & 0.938 & \textbf{0.976} \\
\bottomrule
\end{tabular}
\label{ratios}
\end{table*}

\begin{table*}[h]
\centering
\caption{Experiments on dimension reduction with bottleneck structure. The performance gain is marked \textbf{bold}.}
\begin{tabular}{lcccccccc}
\toprule
\multirow{2}{*}{Dimension reduction?} & \multicolumn{4}{c}{Errors $\downarrow$} & \multicolumn{3}{c}{Errors $\uparrow$} & \multirow{2}{*}{Model size} \\ \cline{2-8} 
 & AbsRel & SqRel & RMS & RMSlog & $<1.25$ & $<1.25^{2}$ & $<1.25^{3}$ \\ \hline
NO & 0.148 & 1.047 & 5.217 & 0.218 & 0.815 & 0.940 & 0.976 & 7.5M \\ 
YES & 0.150 & 1.074 & 5.265 & 0.219 & 0.815 & 0.939 & 0.975 & 5.5M \\ Performance & 1.35\% $\downarrow$ & 2.58\% $\downarrow$ & 0.92\% $\downarrow$ & 0.46\% $\downarrow$ & 0\% $\to$ & 0.11\% $\downarrow$ & 0.10\% $\downarrow$ & \textbf{26.57\% $\uparrow$} \\
\bottomrule
\end{tabular}
\label{bottleneck}
\end{table*}

\begin{table*}[h]
\centering
\caption{Experiments on different number of learning stages. The best performances are marked \textbf{bold}.}
\begin{tabular}{lccccccc}
\toprule
\multirow{2}{*}{\# of learning stages} & \multicolumn{4}{c}{Errors $\downarrow$} & \multicolumn{3}{c}{Errors $\uparrow$} \\ \cline{2-8} 
 & AbsRel & SqRel & RMS & RMSlog & $<1.25$ & $<1.25^{2}$ & $<1.25^{3}$ \\ \hline
5 (c=8) & \textbf{0.150} & 1.074 & \textbf{5.265} & \textbf{0.219} & \textbf{0.815} & \textbf{0.939} & 0.975 \\
4 (c=8) & 0.151 & \textbf{1.065} & 5.273 & \textbf{0.219} & 0.811 & 0.938 & 0.976 \\ 
3 (c=8) & 0.162 & 1.086 & 5.441 & 0.227 & 0.788 & 0.934 & \textbf{0.977} \\ \midrule
5 (c=16) & 0.139 & 0.996 & 5.094 & 0.210 & 0.830 & \textbf{0.948} & \textbf{0.979} \\ 
4 (c=16) & \textbf{0.135} & \textbf{0.973} & \textbf{5.084} & \textbf{0.208} & \textbf{0.840} & \textbf{0.948} & 0.978 \\ 
3 (c=16) & 0.145 & 1.014 & 5.204 & 0.216 & 0.818 & 0.943 & \textbf{0.977} \\ 
\bottomrule
\end{tabular}
\label{stages}
\end{table*}

\begin{table*}[h]
\centering
\caption{Experiments on different padding modes. The best performances are marked \textbf{bold}.}
\begin{tabular}{lccccccc}
\toprule
\multirow{2}{*}{Padding modes} & \multicolumn{4}{c}{Errors $\downarrow$} & \multicolumn{3}{c}{Errors $\uparrow$} \\ \cline{2-8} 
 & AbsRel & SqRel & RMS & RMSlog & $<1.25$ & $<1.25^{2}$ & $<1.25^{3}$ \\ \hline
Replicate & 0.139 & 1.033 & 5.161 & 0.214 & 0.825 & 0.942 & 0.974 \\ 
Zeros & 0.136 & 1.041 & 5.115 & 0.210 & 0.837 & 0.946 & 0.977 \\
Reflect & \textbf{0.135} & \textbf{0.973} & \textbf{5.084} & \textbf{0.208} & \textbf{0.840} & \textbf{0.948} & \textbf{0.978} \\ 
\bottomrule
\end{tabular}
\label{paddingmodes}
\end{table*}

\begin{table*}[h]
\centering
\caption{Experiments on FSLBlock components. The best performances are marked \textbf{bold}.}
\begin{tabular}{lccccccc}
\toprule
\multirow{2}{*}{Blocks} & \multicolumn{4}{c}{Errors $\downarrow$} & \multicolumn{3}{c}{Errors $\uparrow$} \\ \cline{2-8} 
 & AbsRel & SqRel & RMS & RMSlog & $<1.25$ & $<1.25^{2}$ & $<1.25^{3}$ \\ \hline
CNNBlock & 0.143 & 1.061 & 5.208 & 0.215 & 0.826 & 0.943 & 0.977 \\ 
LFLBlock & 0.235 & 1.821 & 6.688 & 0.295 & 0.646 & 0.874 & 0.954 \\
CNNBlock$\to$LFLBlock (in series) & 0.159 & 1.140 & 5.351 & 0.223 & 0.790 & 0.934 & 0.977 \\
LFLBlock$\to$CNNBlock (in series) & 0.150 & 1.082 & 5.234 & 0.218 & 0.810 & 0.939 & 0.977 \\
CNNBlock+LFLBlock (in parallel) & \textbf{0.135} & \textbf{0.973} & \textbf{5.084} & \textbf{0.208} & \textbf{0.840} & \textbf{0.948} & \textbf{0.978} \\ 
\bottomrule
\end{tabular}
\label{localglobal}
\end{table*}

\subsection{Results}

\subsubsection{Self-supervised depth estimation}
Table \ref{kittiresults} presents the quantitative results of depth estimation on the KITTI dataset \cite{geiger2013vision}. Apparently, the proposed FSLNets outperform other state-of-the-art methods by large margins when without pretraining, even outperforming the methods with pretraining in some metrics. Moreover, the proposed models are lightweight; FSLNet-S and FSLNet-L significantly reduce the number of parameters by over $\%78$ and $36\%$ compared with \cite{mypaper2}, respectively. The time complexities are still on a par with other state-of-the-art methods.

As introduced in Section 4, the proposed model is expected to extract the global and local features simultaneously; thus, both global context and fine details in depth maps are expected to be captured. Figure \ref{qualitativeresultonkitti} shows the qualitative comparisons of the proposed model and other state-of-the-art methods, which clearly demonstrate the superiority of the proposed model. More specifically, the reflective regions in the image, such as the cars' windows (the first, second, and fifth columns in Fig. \ref{qualitativeresultonkitti}), are challenging to predict using the local operators; the proposed model can effectively obtain the global context, thereby predicting these difficult regions correctly. Moreover, fine details and far scenes, such as fine fence (the first column in Fig. \ref{qualitativeresultonkitti}) and trees (the third and fourth columns in Fig. \ref{qualitativeresultonkitti}), are also difficult to be fully captured; the proposed model can perform much better than other state-of-the-art methods.

To further evaluate the generalization capability of the proposed model, we directly apply the model trained on the KITTI dataset \cite{geiger2013vision} to the Make3D dataset \cite{saxena2008make3d} without any refinement. Table \ref{make3dresults} reports the quantitative results. When evaluating on the cropped images, the proposed models outperform other state-of-the-art methods by clear margins in the case of without pretraining, even being on a par with those of methods using pretraining. Furthermore, we attempt to resize the entire image (without cropping) to the resolution of $416 \times 128$; then, we predict and evaluate on the entire image. Interestingly, the performance can be further improved in most metrics. We speculate that the strong capability of extracting the comprehensive information (especially extracting the global information) of the proposed model may dominate in this improvement.

Figure \ref{qualitativeresultsonmake3d} presents the qualitative results on the cropped images. Similarly, the proposed model can correctly predict the difficult regions (e.g., large reflective regions), by its excellent capability of capturing the global context and fine details. Figure \ref{qualitativeresultsonmake3dfull} shows some qualitative results on the entire images; the proposed model still can predict a reasonable depth map, though the scene, image resolution, and color distribution significantly vary between the training data and test data.

Table \ref{visualodometry} reports the visual odometry results of the proposed models on the KITTI odometry dataset \cite{geiger2013vision}. The proposed models outperform other state-of-the-art methods by large margins in the condition of without pretraining, even being on a par with the methods adopting pretraining. Moreover, the proposed models, FSLNet-S and FSLNet-L, reduce the space complexities by over $78\%$ and $33\%$ respectively, compared with \cite{mypaper2}.

\subsubsection{Semantic segmentation}
Table \ref{semanticsegmentationresults} presents the semantic segmentation results of the proposed model on the Cityscapes validation dataset \cite{cordts2016cityscapes}. The baseline model (DeepLabv3 \cite{chen2017rethinking}) uses CNNs as the encoder and decoder; under this approach, it takes $580$ epochs to achieve the mIoU of $0.593$ according to \cite{pytorchdeeplibv3}. We substitute the encoder with the proposed FSLNet-L encoder; then, it only takes $340$ epochs to achieve the mIoU of $0.599$, which is slightly better than the results reported in \cite{pytorchdeeplibv3}. Meanwhile, the quick convergence also demonstrates the effectiveness of the proposed model.

Moreover, we apply the proposed FSLNet-S and FSLNet-L; after training of $120$ epochs, FSLNet-S and FSLNet-L attain the mIoUs of $0.625$ and $0.675$, respectively, which outperform the baseline model by large margins. Importantly, the proposed networks significantly reduce the number of parameters and can converge quickly. Figure \ref{qualitativeresultsoncity} shows some qualitative results on the Cityscapes dataset \cite{cordts2016cityscapes}.

\subsection{Ablation studies}
In this subsection, detailed ablation studies are presented. All ablation experiments are conducted on self-supervised depth estimation.

Table \ref{actlayers} studies the different activation functions; the SiLU activation function performs better than other activation functions and therefore is chosen in this work. Table \ref{ratios} studies the different global-local ratios in FSLBlock. Intuitively, it is a kind of tradeoff between global and local information extraction. The more layers of linear frequency learning, the more weights in extracting the global information. By contrast, the fewer layers of convolution, the smaller weight of extracting the local information. Therefore, the target is to search for the best configuration, which enables the network to effectively capture both global and local features. Detailed experiments presented in Table \ref{ratios} indicate that the best proportion of the number of linear frequency learning layers and the number of convolutional layers is $2:4$.

Table \ref{bottleneck} evaluates the bottleneck structure in CNNBlock. The numbers indicate that the bottleneck structure causes a minor performance drop, but significantly reduces the number of parameters of the proposed model. 

Table \ref{stages} experiments with different learning stages. The numbers presented in Table \ref{stages} indicate that the performance of $5$-stage learning is slightly better than $4-$ and $3-$stage learning when $c=8$; while $4-$stage learning significantly outperforms other learning schemes when $c=16$. Thus, we adopt $4$-stage learning and $c \ge 16$.

Table \ref{paddingmodes} experiments with different padding modes. The numbers presented in Table \ref{paddingmodes} indicate that the reflective padding performs slightly better; thus, it is chosen in our networks.

Table \ref{localglobal} conducts the experiments on the components of the proposed FSLBlock. First of all, we solely use the CNNBlock, i.e., the network does not have any global operator. Then, we solely use the proposed LFLBock, excluding CNNBlock; in this context, the network does not have any local operator. The first two rows in Table \ref{localglobal} indicate that both pure global operator and local operator suffer a great performance drop. Moreover, the pure CNNBlock performs much better than the pure LFLBlock, which implies that the local information is more important.

Furthermore, we use both CNNBlock and LFLBlock, but organize them in different structures, i.e., in series and parallel, respectively. The numbers in the last three rows of Table \ref{localglobal} demonstrate that organizing the CNNBlock and LFLBlock in parallel can significantly boost the performance.

\section{LIMITATION \& FUTURE WORK}

There is one issue of the proposed model: relatively high time complexity. As shown in Table \ref{kittiresults}, the time complexity of the proposed FSLNet-L is relatively higher than other models. The causes are as follows.

\begin{itemize}
\item We adopt CNNBlock to capture the local details of the input image, but convolution is a kind of dense computation. In particular, the computational cost will dramatically increase when the overlapping exists between the receptive fields. However, we intend to design the overlapping between the receptive fields for capturing the fine details. Therefore, it comes to be a tradeoff between the computational cost and fine details extraction in CNNBlock.

\item We rely on Fourier transform (FT) to capture the global information of the input image, but the time complexity of FT is $O(nlogn)$, which is relatively high. Moreover, we perform several times of FT and IFT in the proposed FSLNet to integrate the global and local information, which raises the computational cost.

\end{itemize}

Thus, we will focus on improving the issues mentioned above in the future.

\section{CONCLUSION}

Feature extraction and transformation play a critical role in deep learning. However, the current learning process mainly focuses on spatial domain learning, current models of which are incapable to capture both global and local information simultaneously. To this end, we propose joint learning of frequency and spatial domains, taking full advantage of frequency learning and spatial learning to capture the global context and local details, respectively.

Specifically, we propose the linear frequency learning block (LFLBlock), the joint frequency and spatial domains learning block (FSLBlock), and the joint frequency and spatial domains learning network (FSLNet), respectively. The proposed model is applied to two dense predictions tasks, i.e., self-supervised depth estimation (including one sparse prediction task, ego-motion estimation) and semantic segmentation. Detailed experiments demonstrate the effectiveness and superiority of the proposed model. We hope that the presented work contributes to related industrial and academic communities, as well as encourages more research focusing on cross-domain learning.

% or


\appendix[Feature map visualization]

\begin{figure*}[h]
    \centering
    \begin{overpic}[width=\textwidth]{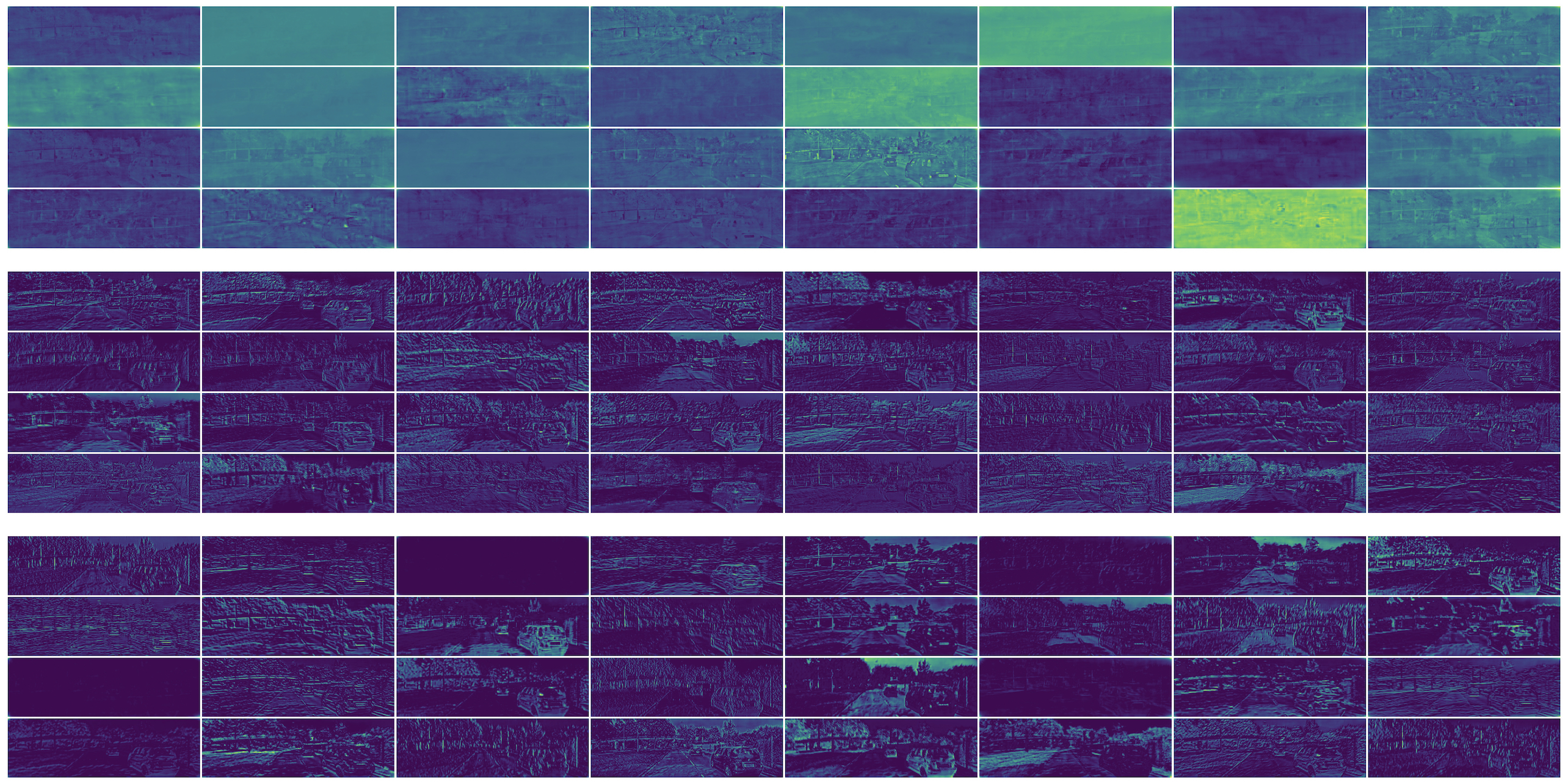}
    \put(-1,3){\scriptsize\rotatebox{90}{Fusion features}}
    \put(-1,21){\scriptsize\rotatebox{90}{Local features}}
    \put(-1,37){\scriptsize\rotatebox{90}{Global features}}
    \end{overpic}
    \caption{Feature map visualization of FSLBlock1 of FSLNet-L. Global features are obtained from LFLBlock; local features are obtained from CNNBlock; fusion features are obtained from the fusion of global features and local features. The input image is the first one in Fig. \ref{qualitativeresultonkitti}.}
    \label{f0}
\end{figure*}

\begin{figure*}[h]
    \centering
    \begin{overpic}[width=\textwidth]{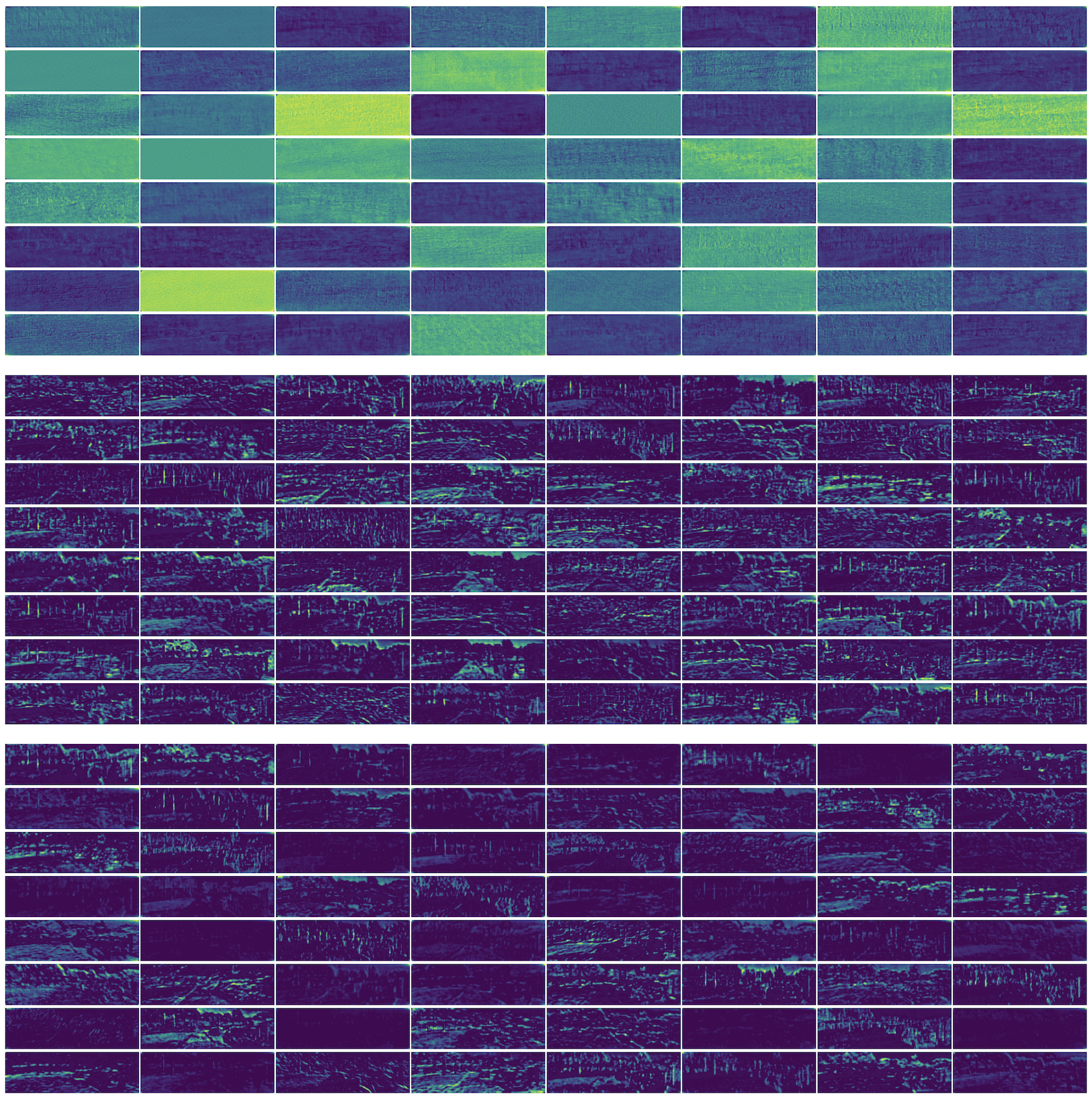}
    \put(-1,10){\scriptsize\rotatebox{90}{Fusion features}}
    \put(-1,45){\scriptsize\rotatebox{90}{Local features}}
    \put(-1,80){\scriptsize\rotatebox{90}{Global features}}
    \end{overpic}
    \caption{Feature map visualization of FSLBlock2 of FSLNet-L. Global features are obtained from LFLBlock; local features are obtained from CNNBlock; fusion features are obtained from the fusion of global features and local features. The input image is the first one in Fig. \ref{qualitativeresultonkitti}.}
    \label{f1}
\end{figure*}

\begin{figure*}[h]
    \centering
    \begin{overpic}[height=\textheight]{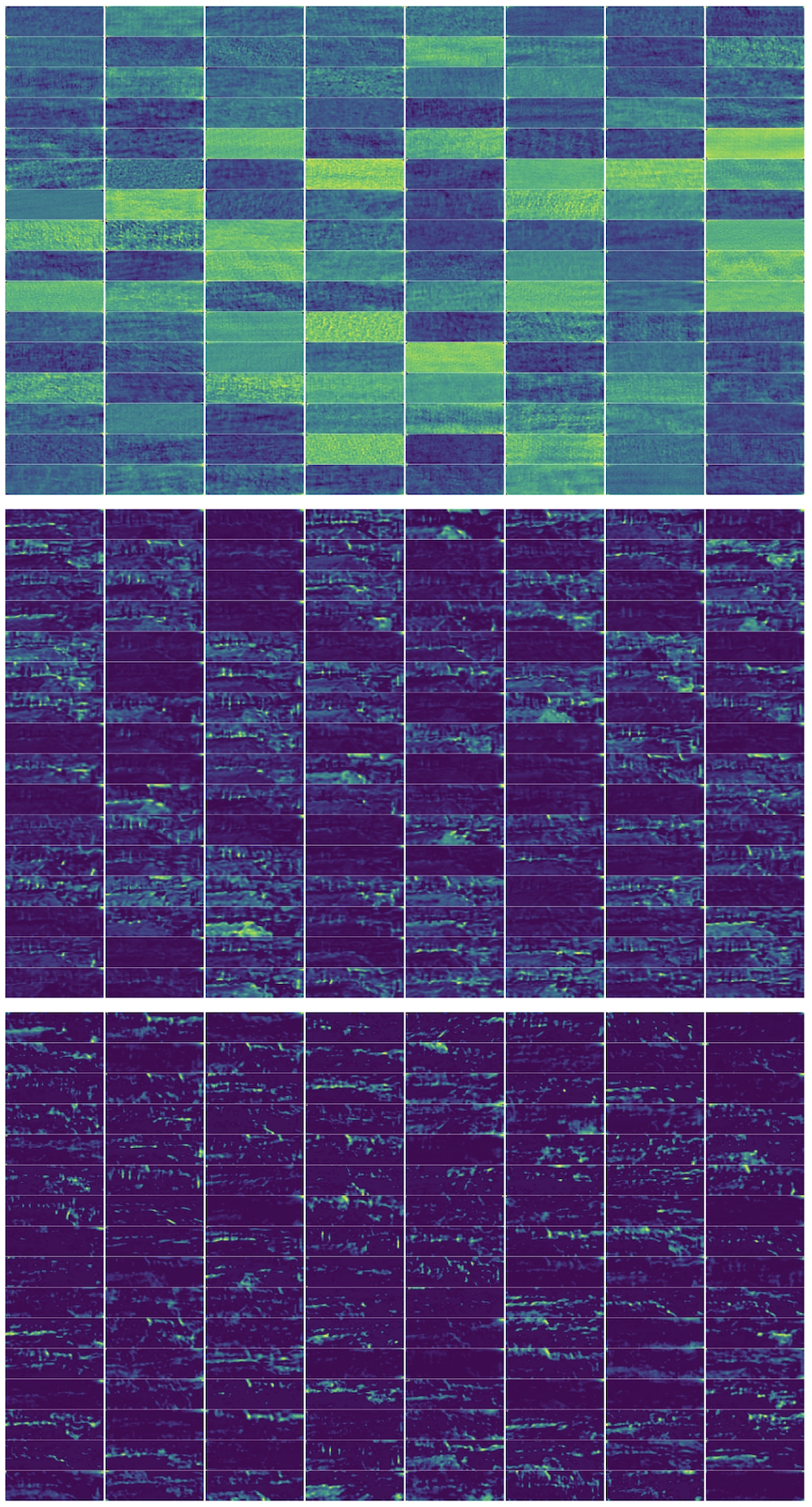}
    \put(-1,12){\scriptsize\rotatebox{90}{Fusion features}}
    \put(-1,47){\scriptsize\rotatebox{90}{Local features}}
    \put(-1,80){\scriptsize\rotatebox{90}{Global features}}
    \end{overpic}
    \caption{Feature map visualization of FSLBlock3 of FSLNet-L. Global features are obtained from LFLBlock; local features are obtained from CNNBlock; fusion features are obtained from the fusion of global features and local features. The input image is the first one in Fig. \ref{qualitativeresultonkitti}.}
    \label{f2}
\end{figure*}

\begin{figure*}[h]
    \centering
    \begin{overpic}[width=\textwidth]{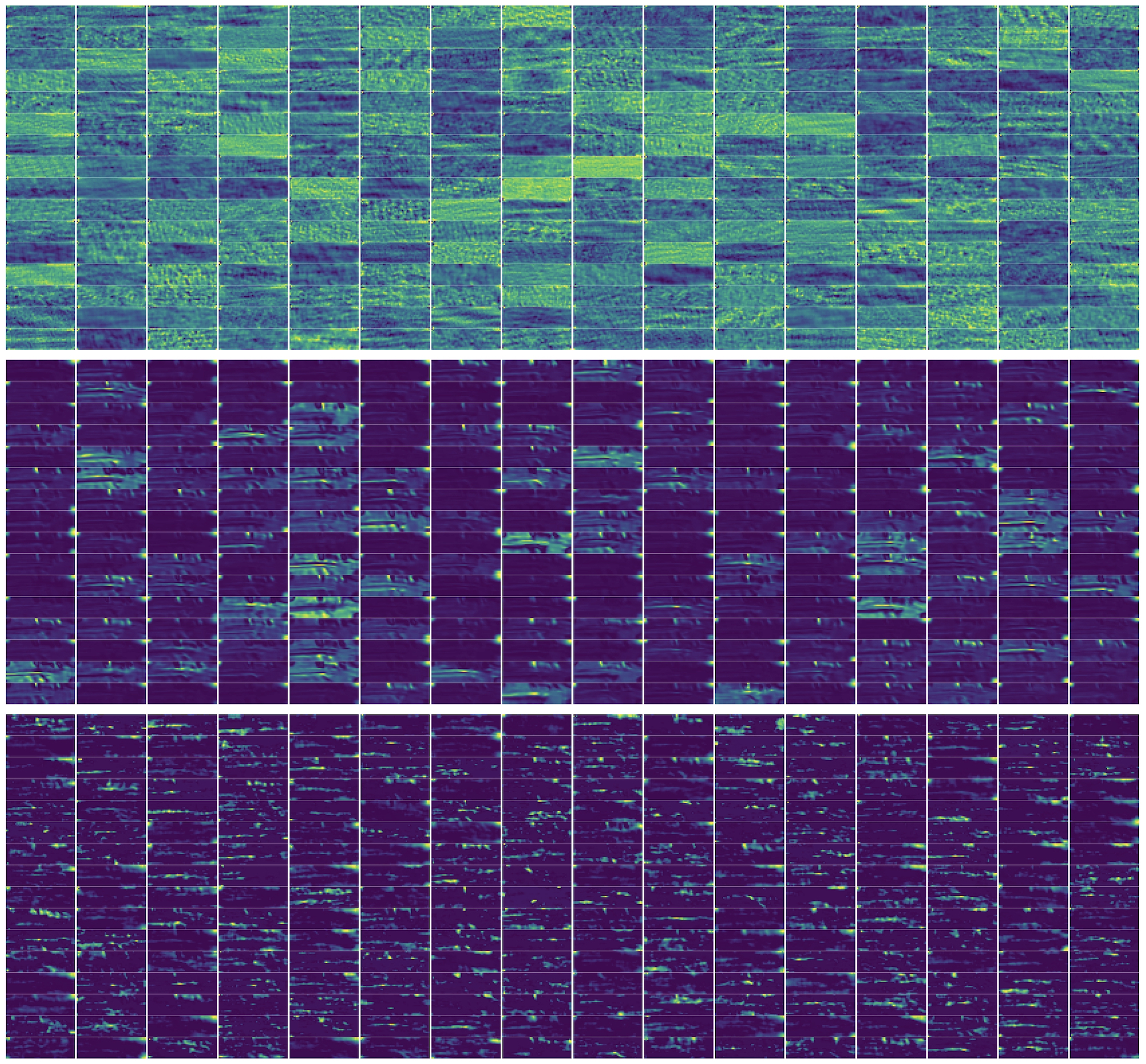}
    \put(-1,12){\scriptsize\rotatebox{90}{Fusion features}}
    \put(-1,44){\scriptsize\rotatebox{90}{Local features}}
    \put(-1,75){\scriptsize\rotatebox{90}{Global features}}
    \end{overpic}
    \caption{Feature map visualization of FSLBlock4 of FSLNet-L. Global features are obtained from LFLBlock; local features are obtained from CNNBlock; fusion features are obtained from the fusion of global features and local features. The input image is the first one in Fig. \ref{qualitativeresultonkitti}.}
    \label{f3}
\end{figure*}

\begin{figure*}[h]
    \centering
    \begin{overpic}[height=\textheight]{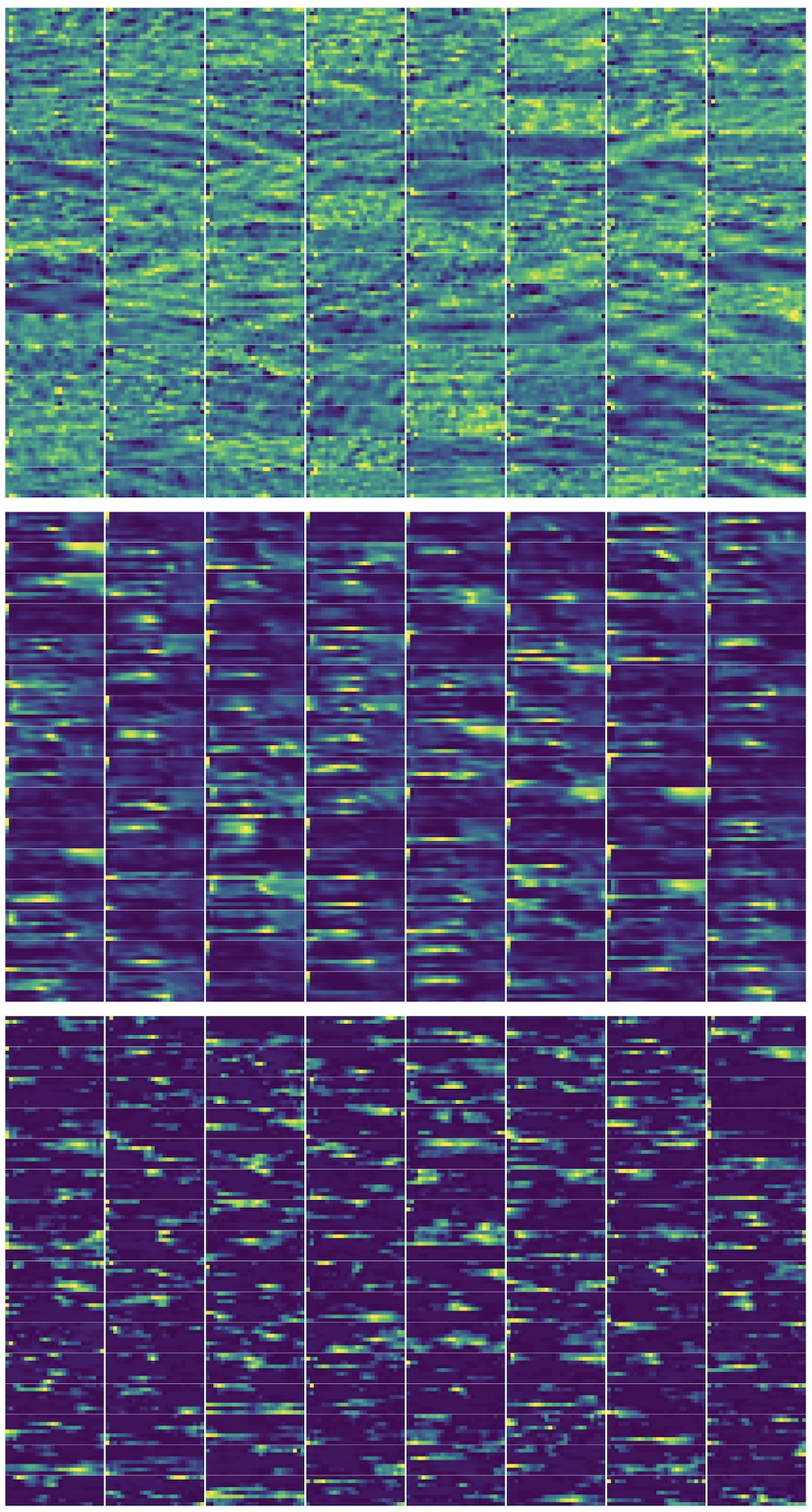}
    \put(-1,12){\scriptsize\rotatebox{90}{Fusion features}}
    \put(-1,47){\scriptsize\rotatebox{90}{Local features}}
    \put(-1,80){\scriptsize\rotatebox{90}{Global features}}
    \end{overpic}
    \caption{Feature map visualization of FSLBlock5 of FSLNet-L. Global features are obtained from LFLBlock; local features are obtained from CNNBlock; fusion features are obtained from the fusion of global features and local features. The input image is the first one in Fig. \ref{qualitativeresultonkitti}.}
    \label{f4}
\end{figure*}

\begin{figure*}[h]
    \centering
    \begin{overpic}[width=\textwidth]{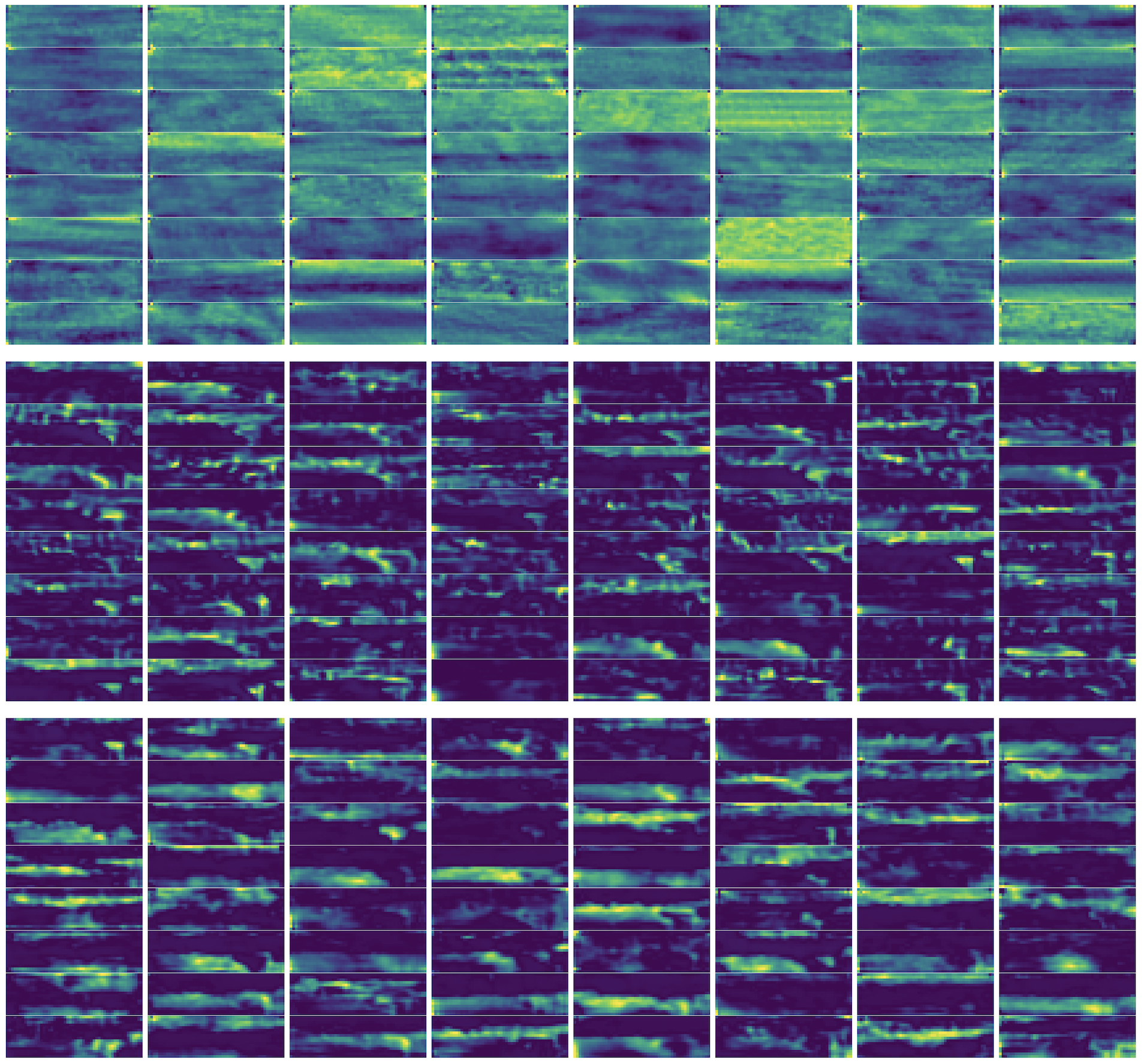}
    \put(-1,10){\scriptsize\rotatebox{90}{Fusion features}}
    \put(-1,45){\scriptsize\rotatebox{90}{Local features}}
    \put(-1,80){\scriptsize\rotatebox{90}{Global features}}
    \end{overpic}
    \caption{Feature map visualization of FSLBlock6 of FSLNet-L. Global features are obtained from LFLBlock; local features are obtained from CNNBlock; fusion features are obtained from the fusion of global features and local features. The input image is the first one in Fig. \ref{qualitativeresultonkitti}.}
    \label{f5}
\end{figure*}

\begin{figure*}[h]
    \centering
    \begin{overpic}[width=\textwidth]{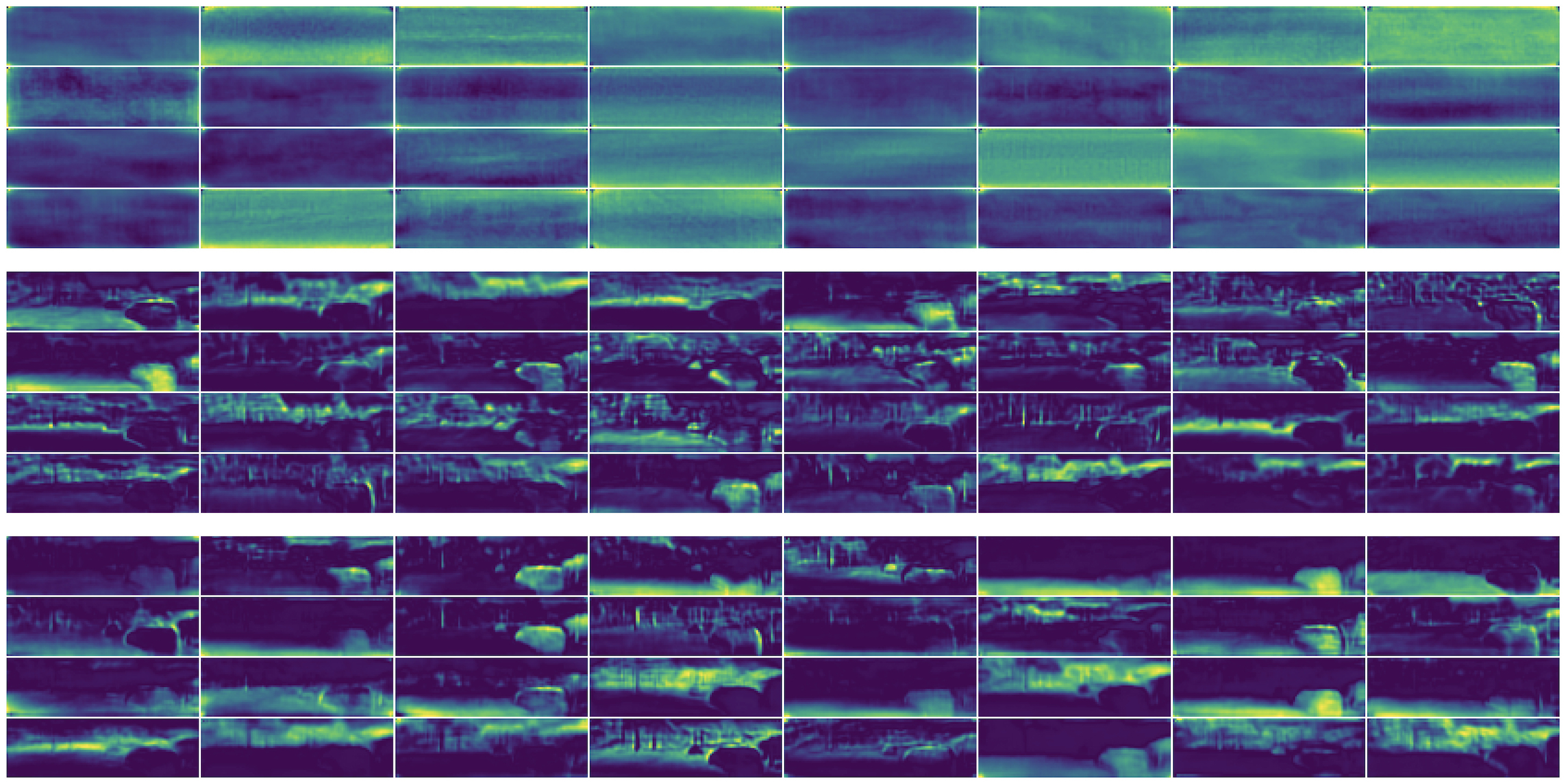}
    \put(-1,3){\scriptsize\rotatebox{90}{Fusion features}}
    \put(-1,21){\scriptsize\rotatebox{90}{Local features}}
    \put(-1,37){\scriptsize\rotatebox{90}{Global features}}
    \end{overpic}
    \caption{Feature map visualization of FSLBlock7 of FSLNet-L. Global features are obtained from LFLBlock; local features are obtained from CNNBlock; fusion features are obtained from the fusion of global features and local features. The input image is the first one in Fig. \ref{qualitativeresultonkitti}.}
    \label{f6}
\end{figure*}

\begin{figure*}[h]
    \centering
    \begin{overpic}[width=\textwidth]{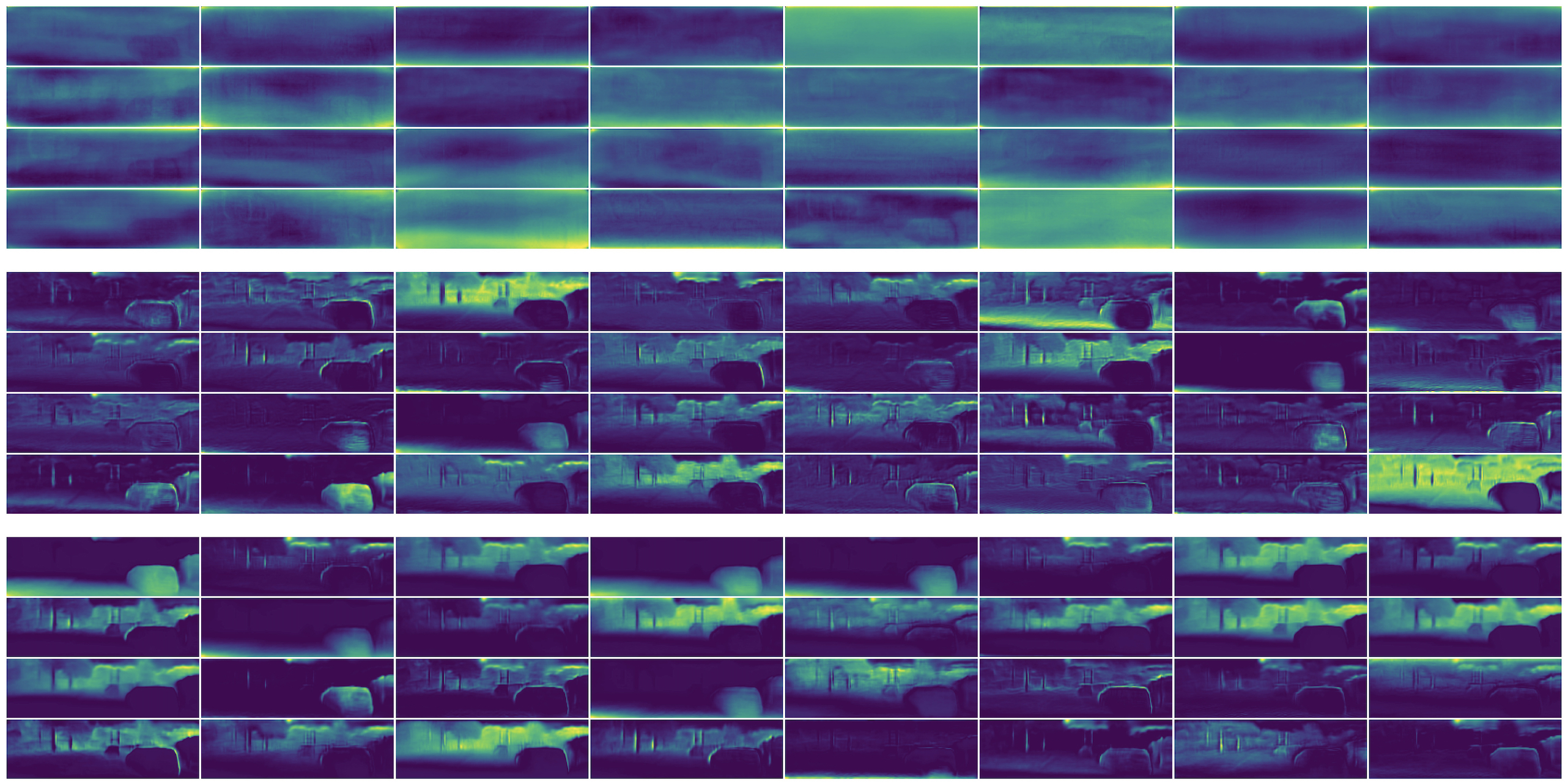}
    \put(-1,3){\scriptsize\rotatebox{90}{Fusion features}}
    \put(-1,21){\scriptsize\rotatebox{90}{Local features}}
    \put(-1,37){\scriptsize\rotatebox{90}{Global features}}
    \end{overpic}
    \caption{Feature map visualization of FSLBlock8 of FSLNet-L. Global features are obtained from LFLBlock; local features are obtained from CNNBlock; fusion features are obtained from the fusion of global features and local features. The input image is the first one in Fig. \ref{qualitativeresultonkitti}.}
    \label{f7}
\end{figure*}

To visualize the feature maps, we shift the feature maps to the range of 0 to 1, by subtracting the minimum and dividing by the maximum value. Figures \ref{f0} to \ref{f7} show the visualized feature maps; each figure is divided into three cells, i.e., global features, local features, and fusion features. The visualization results perfectly meet the expectations. Specifically, the proposed LFLBlock can capture global information, but has difficulty in capturing local information; in this context, the output of LFLBlock should be smooth features. By contrast, CNNBLock can capture local information very well; thus, the output of CNNBlock should contain fine details, but may not sense the global context very well. Fusion strategy, which combines global and local information, can obtain comprehensive features, predicting the result much better than any single type of feature.

%\section{}

%\ifCLASSOPTIONcompsoc
%\else
%\fi


\ifCLASSOPTIONcaptionsoff
  \newpage
\fi



%\bibliographystyle{IEEEtran}
\bibliography{IEEEabrv,ref.bib}{}
\bibliographystyle{IEEEtran}
%\begin{thebibliography}{1}


%\end{thebibliography}


%\end{IEEEbiography}


%\newpage



%\vfill


\end{document}